\documentclass[10pt]{article} 
\usepackage{soul} 
\usepackage{color, xcolor} 

\usepackage[preprint]{tmlr}
\usepackage{makecell}
\usepackage{hyperref}
\usepackage{url}
\usepackage{times}
\usepackage{latexsym}
\usepackage{graphicx}
\usepackage{inconsolata}
\usepackage{graphicx}
\usepackage{amsmath}
\usepackage{booktabs}
\usepackage{arydshln}
\usepackage{multirow}
\usepackage{wrapfig, lipsum, booktabs}
\usepackage{amsfonts}
\usepackage{algorithm}
\usepackage{enumerate}
\usepackage{enumitem}
\usepackage{cases}
\usepackage[T1]{fontenc}
\usepackage{fdsymbol}
\usepackage{CJK}
\usepackage{indentfirst}
\usepackage{amsmath}
\usepackage{cases}
\usepackage{algorithmic}
\usepackage{setspace}

\usepackage[utf8]{inputenc}

\usepackage{microtype}

\usepackage{inconsolata}
\title{
\begin{center}
    Venturing into Uncharted Waters: The Navigation Compass from Transformer to Mamba
\end{center}}

\author{
\begin{center}\name Yuchen Zou, ~Yineng Chen, ~Zuchao Li \thanks{Correspondence to: Zuchao Li <zcli-charlie@whu.edu.cn>.}, ~~Lefei Zhang, ~Hai Zhao
\end{center}
}

\def\month{MM}  
\def\year{YYYY} 
\def\openreview{\url{https://openreview.net/forum?id=XXXX}} 

\begin{document}
\maketitle

\begin{figure*}[htb]
    \centering
    \vspace{-3mm}
    \includegraphics[width=1.0\textwidth]{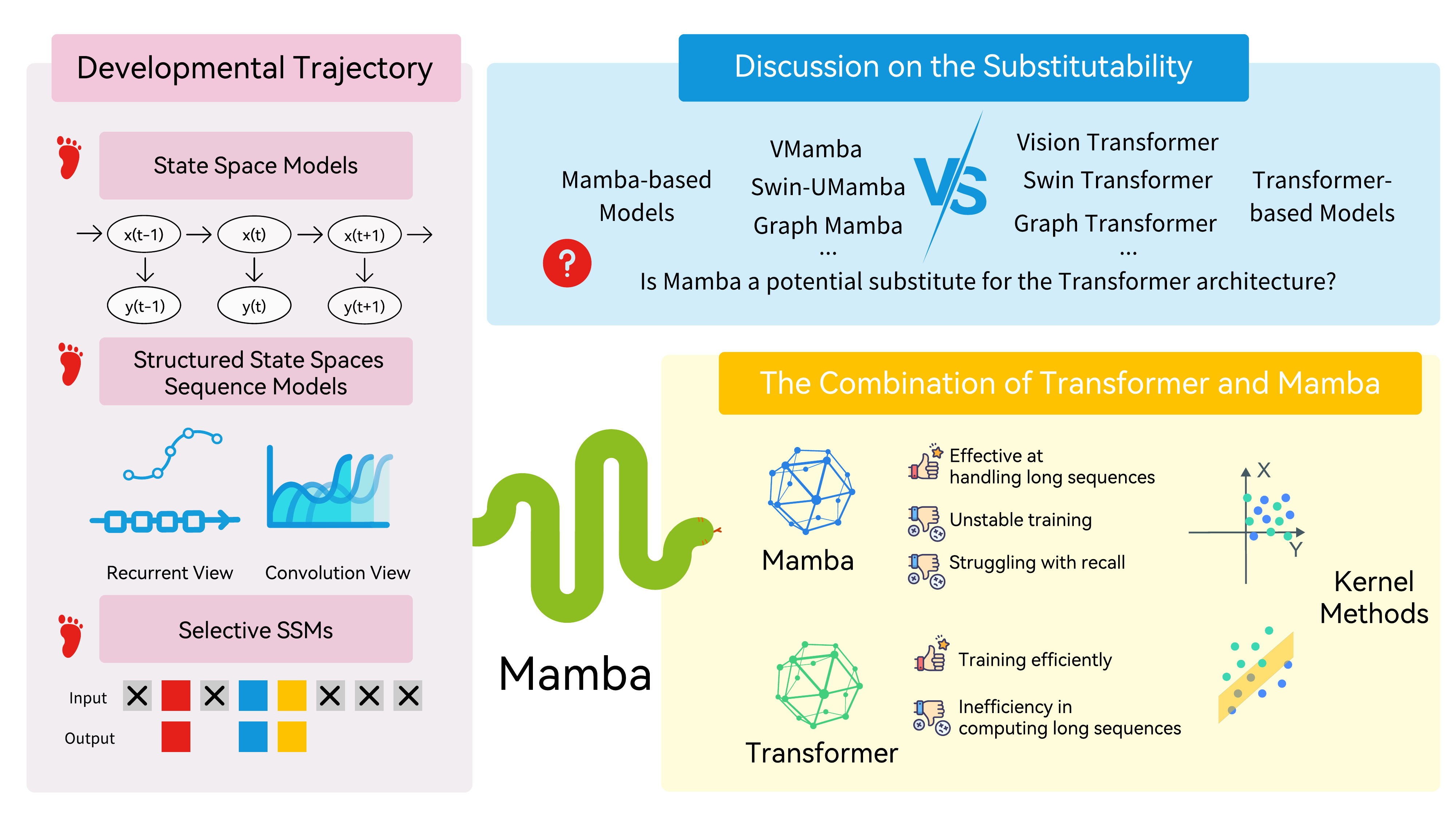}
    \vspace{-6mm}
    \caption{An overview of Mamba, including the developmental trajectory of structured state space models, a discussion on the substitutability of Mamba and Transformers, and their combination.}
    \label{fig:cot-role}
    \vspace{-1mm}
\end{figure*}

\begin{abstract}
Transformer, a deep neural network architecture, has long dominated the field of natural language processing and beyond. Nevertheless, the recent introduction of Mamba challenges its supremacy, sparks considerable interest among researchers, and gives rise to a series of Mamba-based models that have exhibited notable potential. This survey paper orchestrates a comprehensive discussion, diving into essential research dimensions, covering: (i) the functioning of the Mamba mechanism and its foundation on the principles of structured state space models; (ii) the proposed improvements and the integration of Mamba with various networks, exploring its potential as a substitute for Transformers; (iii) the combination of Transformers and Mamba to compensate for each other's shortcomings. We have also made efforts to interpret Mamba and Transformer in the framework of kernel functions, allowing for a comparison of their mathematical nature within a unified context. Our paper encompasses the vast majority of improvements related to Mamba to date.

\end{abstract}
\footnotetext[1]{This work was supported by the National Natural Science Foundation of China (No.62306216),
the Natural Science Foundation of Hubei Province of China
(No.2023AFB816), the Fundamental Research Funds for the Central Universities (No.2042023kf0133).}

\section{Introduction}
Since the introduction of the transformer \citep{vaswani2017attention} architecture, which surpassed previous state-of-the-art recurrent neural network models in translation tasks, numerous large language models (LLMs) built upon the transformer framework such as GPT \citep{radford2018improving} and BERT \citep{devlin2018bert} have emerged. Moreover, Transformer has been a prevalent deep learning model extensively applied across diverse domains besides natural language processing and computer vision. However, the drawbacks of the transformer primarily lie in two aspects: firstly, its inefficiency in computing long sequences, and secondly, the limitation of its core attention module, which models only within a finite context window, and this results in performance deficiencies when handling longer sequences.

Based on recent studies on state space models \citep{fu2022hungry,poli2023hyena} and previous studies on structured state space sequence models (S4) \citep{gu2021efficiently,gu2021combining}, Mamba architecture has emerged with the expectation of replacing attention modules structurally. Mamba also introduces a selection mechanism that makes the SSM parameters a function of the input, achieving selective propagation and information-forgetting effects. In other words, the model can focus on or disregard specific inputs, filtering out irrelevant information and maintaining long-term memory of information relevant to the problem. Additionally, Mamba incorporates a hardware-aware algorithm, opting for a scanning approach instead of convolution for model computations, in contrast to the strategy adopted by all previous SSM models for computational efficiency, which relied on fixed time and input. This algorithm is akin to Flash Attention \citep{dao2022flashattention,dao2023flashattention}, both aimed at addressing the slow and memory-hungry nature of Transformers when dealing with long sequences with a focus on hardware-level solutions.

Due to its remarkable achievements that Mamba exhibits 5× generation throughput compared with Transformer of similar size and Mamba-3B demonstrates comparable performance to Transformer twice its size, there has been a recent surge in models applying Mamba to  various domains, including image haze removal \citep{zheng2024u}, table data processing \citep{ahamed2024mambatab}, CT-MR image manipulation \citep{guo2024mambamorph}. Researchers have also combined Mamba with various models such as Mixture of Experts (MoE) \citep{pioro2024moe, anthony2024blackmamba}, GNNs \citep{behrouz2024graphmamba, wang2024graph-mamba}, U-Net \citep{ma2024u}, further expanding its applicability. Nevertheless, whether Mamba can truly replace the Transformer remains a question of heated discussion.

In this survey, we navigate through research topics that cover: (i) the operation of the Mamba mechanism, grounded in the principles of structured state space models; (ii) the substitutability of Mamba for models including Vison Transformer (ViT), Graph Transformer (GT), Swin Transformer and U-Net; (iii) the combination of Transformers and Mamba to compensate for each other’s shortcomings. In the rest of this paper, We have also tried to interpret Mamba and Transformer using kernel functions, allowing us to compare their mathematical characteristics within a single framework. In the subsequent section, we explore the extent to which Mamba can potentially replace Transformer, examining the domains in which such substitution may occur and identifying any limitations. To conclude, we also discuss its current advantages and disadvantages.

This paper goes beyond a simple summary, aiming to explore and elucidate the essence of Mamba, as well as discussing its potential as a substitute for Transformer. Our carefully selected papers delve into the underlying preliminaries of Mamba\citep{kalman1960new,koller2009probabilistic,gu2021combining,gu2021efficiently,gu2020hippo,gu2022train}, the mathematical principles explaining Mamba and Transformer \citep{gu2022train}, and Mamba-based models\citep{zhu2024vision,liu2024vmamba}. Additionally, our paper documents the developmental trajectory of state space models, providing a comprehensive overview of the evolution of state space models and highlighting both differences and connections.

\vspace{-2.5mm}
\paragraph{Key Takeaways} 
We systematically and comprehensively elucidate the underlying mechanisms of Mamba. The key takeaways are:

\begin{itemize}[leftmargin=*,topsep=2pt,itemsep=1pt,parsep=0pt]
    \item An overview of the developmental trajectory of state space-related models, tracing the evolution from SSMs to structured state space sequence models to selective SSMs.  (Section \ref{develop}).
    \item A discussion on the substitutability of Mamba-based models for Transformer-based models (Section \ref{sec:substitute}).
    \item A discussion on the combination of Transformers and Mamba and a kernel method explanation of the mathematical essence including  structures of Transformer and
Mamba-based model.(Section \ref{combine}). 
    \item Despite the achievements of Mamba, its limitations and challenges are acknowledged and discussed (Section \ref{conclude}).
\end{itemize}

\section{Development of State Space Models}
\label{develop}
This section delves into the fundamental mechanisms of Mamba, tracing its developmental history from the traditional state space models  to structured state space sequence models and further to Selective SSM (S6).

\subsection{State Spaces Models}
\label{sec:ssm}
State space model (SSM) refers to a class of probabilistic graphical models \citep{koller2009probabilistic} in the field of machine learning. The term "state space" was initially introduced by \citeauthor{kalman1960new} and was originally applied in the field of control engineering. Dean and Kanazawa introduced the initial temporal extension of probabilistic graphical models \citep{dean1989model}. State space models come in various types \citep{durbin2012time}, including linear state space models, nonlinear state space models, non-Gaussian state space models, and others. The objective of SSM is to compute the optimal estimation, i.e., the output, based on observed input. SSM is based on three variables that evolve: input $\{u_{1}, \ldots, u_{n}\}$, state $\{x_{1}, \ldots, x_{n}\}$, and output $\{y_{1}, \ldots, y_{n}\}$. $x_{i}$ represents latent variables, which means unobserved variables while $u_{i}$ corresponds to observed state variables. $y_{i}$ is deduced based on the probabilistic dependency between these two variables. 




For the sake of simplicity, let time be denoted as $t$, the input as $u(t)$, the state as $x(t)$, and $y(t)$ represent the output. The most general state-space representation of a linear system is expressed as follows \citep{brogan1991modern}:

\begin{subequations}
\begin{align}
x'(t) &= A(t)x(t)+B(t)u(t) \label{x'(t)}\\ 
y(t) &= C(t)x(t)+D(t)u(t)    \label{y(t)}     
\end{align}\label{classic}
\end{subequations}

where $A$ is the state matrix, $B$ is the input matrix, $C$ is the output matrix, and $D$ is the command matrix. We use $x'(t)$ to denote the derivative of x(t).






\subsection{Structured State Spaces Sequence Models}
\label{sec:s4}
Structured State Spaces Sequence Models (S4) \citep{gu2021efficiently} were proposed based on a novel parameterization for the SSM, designed to handle data containing long-range dependencies (LRDs) efficiently. S4 can be viewed as a combination of SSM, HiPPO, and structured matrices. Under the assumption of linear time invariance (LTI), the matrices will remain constant. Moreover, in the context of deep learning, equation (\ref{y(t)}) can be further simplified by  equation (\ref{ssm2}), where $Du(t)=0$ is considered as a skip connection and it can be computed easily. 

\begin{subequations}
\begin{align}
x'(t) &= Ax(t)+Bu(t) \label{ssm1}\\ 
y(t) &= Cx(t)     \label{ssm2}     
\end{align}\label{ssmeq}
\end{subequations}

\subsubsection{Discretization}
The continuous view can handle continuous data but it is rather slow both in training and inference \citep{gu2021combining}. If we aim to apply SSM on a discrete input sequence, it is necessary to first discretize the parameters \citep{gu2022train}. One of the main differences between various S4 architectures arises from the discrepant methods of discretization. There are two distinct approaches here: the recurrent view and the convolution view. However, mathematically, they are equivalent. 
\paragraph{The Recurrent View.}
Taking the commonly used Euler discretization method as an example, applying a differential approach:

\begin{align}
x(t+\Delta) &\approx x(t)+\Delta x'(t) \notag \\
            &=x(t)+\Delta\left(Ax(t)+Bu(t)\right)  \\
            &=\left(I+\Delta A\right)x(t)+\Delta Bu(t)\notag \\
            &=\overline{A}x(t)+\overline{B}u(t) \notag
\end{align}

As a result, we transform the continuous-time model into a discrete-time model with cyclic updates. The formula (\ref{ssmeq}) can be unrolled as a linear RNN. The Recurrent View provides the benefits of unbounded context in principle and highly efficient inference.

\begin{subequations}
\begin{align}
x_k &=\overline{A}x_{k-1}+\overline{B}u_k \label{ssm1rec}\\ 
y_k &= \overline{C}x_k     \label{ssm2rec}     
\end{align}\label{ssmrec}
\end{subequations}

\paragraph{The Convolution View.}
Assuming the input $u_i$ is known, starting from $x_0$ and $y_0$, we can sequentially compute all $x_i$ and $y_i$. Then equation (\ref{ssmrec})can be explicitly expanded in closed form, yielding the following result:

\begin{minipage}{0.25\textwidth}
\begin{subequations}
\begin{align}
x_0 &=\overline{B}u_0 \notag\\ 
y_0 &= \overline{CB}u_0 \notag      
\end{align}
\end{subequations}
\end{minipage}
\begin{minipage}{0.25\textwidth}
\begin{subequations}
\begin{align}
x_1 &=\overline{AB}u_0+\overline{B}u_1 \notag\\ 
y_1 &= \overline{CAB}u_0+\overline{CB}u_1  \notag     
\end{align}
\end{subequations}
\end{minipage}
\begin{minipage}{0.5\textwidth}
\begin{subequations}
\begin{align}
x_2 &={\overline{A}}^2\overline{B}u_0+\overline{AB}u_1+\overline{B}u_2 \quad \quad &\cdots \notag \\  
y_2 &={\overline{CA}}^2\overline{B}u_0+\overline{CAB}u_1+\overline{CB}u_2 \quad &\cdots       \notag
\end{align}
\end{subequations}
\end{minipage}
\vspace{2mm}

Continuing in this manner, the general form of $y_k$ is as follows:

\begin{equation}
    y_k ={\overline{CA}}^k\overline{B}u_0+{\overline{CA}}^{k-1}\overline{B}u_1+\cdots+\overline{CAB}u_{k-1}+\overline{CB}u_k
\end{equation}
\begin{equation}
    \overline{K} \in \mathbb{R}^L :=(\overline{CB},\overline{CAB},\cdots,{\overline{CA}}^{L-1}\overline{B})
\end{equation}

\vspace{0.5mm}
In summary, equation (\ref{con}) resembles a CNN and shares a similar structure to convolution. The Convolution View offers the advantages of local information and parallelizable training.

\begin{equation}
    y=\overline{K}*u \label{con}
\end{equation} 
\vspace{2mm}

\subsubsection{HiPPO}
High-order Polynomial Projection Operator (HiPPO) \citep{gu2020hippo} is a general framework for online memorization for long context. The computational efficiency challenge in dealing with long sequence data primarily lies in compressing the history of input within the memory budget to learn features. Typically, the idea of online function approximation is employed and orthogonal polynomials are used as basis. When more signals are received, there is a desire to compress the entire signal still within the memory budget. In contrast to the approach of finding the optimal projection $g(t)$ of input $f$ onto the space of polynomials, which requires complex inner product computations, using coefficient updates to represent the past states over some time, the derivation of \citeauthor{gu2020hippo} suggests that obtaining the coefficients  $c(t)$  of the polynomial can be easily achieved online through solving ordinary differential equations (ODEs). In other words, updating the coefficients $c(t)$ of the polynomial is done such that it provides the optimal approximation of input function $f(t)$, throughout the entire duration of the input from start to end. In this context, $c(t)$ represents the hidden state mentioned in Section \ref{sec:ssm}.
For certain $A(t)\in \mathbb{R}^{N\times N}$, $B(t)\in \mathbb{R}^{N\times 1}$, $c(t)$ takes the form of a simple linear differential equation: 

\begin{equation}
    \frac{d}{dt}c(t)=A(t)c(t)+B(t)f(t)
\end{equation}

Due to the inherently discrete nature of the data, performing discrete operations leads to the discrete-time HIPPO recurrence:

\begin{equation}
   c_{k+1}=A_{k+1}c_{k}+B_{k+1}f_{k} \label{ck}
\end{equation}

In the context of the SSM, the crucial aspect is utilizing the HiPPO operator to choose an appropriate $A$ and $B$ matrix and reconstruct the entire input $u(t)$ with the best fit based on the current state $x(t)$. The matrix that computed by HiPPO and used in S4 seems to have no known mathematical interpretation and was originally defined for time-variant dynamical systems \citep{patro2024mamba360}. The instantiations of HiPPO depend on the selection of the measure, i.e., the weight that expresses the importance of different parts of the history input. For example, the HiPPO-LegT is the built on the measure of the uniform distribution on the most recent history interval, and the HiPPO-LagT adopts exponentially decaying measure. \citeauthor{gu2020hippo} also proposes LegS, a novel measure that scales the uniform distribution to the whole history dynamically. This means that HiPPO-LegS can dynamically adjust weights, ensuring that historical data from different periods are appropriately considered. The matrix form of the HiPPO-LegS is as below:
\begin{numcases}{A_{nk}=}
 (2n+1)^{1/2}(2k+1)^{1/2}  &$n>k$ \notag \\
 k+1 &$n=k$ \\
 0 &$n<k$ \notag
\end{numcases}

To further simplify the computation, \citeauthor{gu2020hippo} proposed S4D, removing the low-rank portion of the HiPPO-LegS. The detailed differences between this simplified model and S4 are as below:
\begin{itemize}[leftmargin=*,topsep=2pt,itemsep=1pt,parsep=0pt]
\item \textbf{Simplified Algorithm}: Compared to the sophisticated algebraic techniques of S4, S4D utilizes Vandermonde matrix multiplication to compute the kernel.
\item \textbf{Simplified structure}: The diagonal state can be viewed as a collection of one-dimensional state space models, which facilitates understanding and implementation.
\item \textbf{Based on Complex Matrices}: The matrices are diagonalized on the complex plane. In addition, the real part is forced to be negative to ensure the stability.
\item \textbf{Performance Comparison}: Most experiments demonstrate S4D's comparable performance to S4, which suggests that S4D offers a simpler implementation choice without sacrificing performance.
\end{itemize}

Based on the above idea, S4D-Lin is proposed as a replacement of S4D-LegS and it is actually one of the initialization scheme of Mamba. 
\begin{equation}
    A_n=-\frac{1}{2}+i\pi n
\end{equation}
The other scheme used in Mamba is S4D-Real,
\begin{equation}
    A_n=-(n+1)
\end{equation}
However, according to \citeauthor{gu2020hippo}, S4D-Real has the same spectrum as S4-LegS, thus its performance is inferior to S4D-Lin.
\subsection{Selective SSMs}
\label{s6}
\citeauthor{gu2023mamba} argue that a basic problem of sequence modeling is compressing context into a smaller state, retaining useful information at the same time. Based on S4, Selective SSMs (S6) were proposed with the core idea of making parameters dependent on the input rather than being time-invariant. However, this introduced some challenges, such as the problem of computational efficiency. The proposed hardware-aware algorithm effectively addresses this issue. 

\subsubsection{Selection Mechanism}
\label{selection}
Many previous Large Language Models (LLMs) can be easily distracted by irrelevant context and do not improve their performance with increasing context length \citep{shi2023large}. However, the model can focus on inputs of interest and filter out potentially irrelevant noise with the adoption of the selection mechanism. Additionally, performance improves by simply resetting the current state to forget the historical records. LTI models cannot separate multiple sequences while the selective mechanism addresses this issue by resetting the state when the model completes processing one sequence and starts handling the next one.

S6 defines several input parameters, $\Delta$, $B$, and $C$, and L functions with an additional dimension for the length of the data. $\Delta$ can be seen as the RNN gating mechanism, representing a step size used for discretization, and the input $u_k$ can be considered as a sampling of the underlying continuous signal $u(t)$, where $u_k=u(k\Delta)$. A larger $\Delta$ places more emphasis on the current input, in other words, it selects the current input and forgets the current state. On the contrary, a smaller $\Delta$ tends to overlook the current input and maintains the current state. S6 performs discretization to transform continuous data $\Delta$,  $A$, $B$, into discrete data $\overline{A}$, $\overline{B}$. $A$ employs Zero-Order Hold (ZOH) discretization, while $B$ uses Euler discretization. 
\begin{equation}
    \overline{A}=exp(\Delta A)
\end{equation}
\begin{equation}
    \overline{B}=(\Delta A)^{-1}(exp(\Delta A)-I)\cdot\Delta B
\end{equation}

\paragraph{The explanation of why ZOH discretization is used.}
The state space model is actually a one-order linear differential equation:
\begin{equation}
    \frac{dx(t)}{dt}=Ax(t)+Bu(t)
\end{equation}

After multiplying the integrating factor $e^{\int Adt}$, it is easy to get the solution
\begin{equation}
x(t)=e^{\int_{t_0}^tAd\gamma}x(t_0)+\int_{t_0}^t Be^{\int_{s}^tAd\gamma} u(s) ds
\end{equation}
Set $t=t_{k+1}$ and $t_0=t_{k}$, we have
\begin{equation}
    x(t_{k+1})=e^{A(t_{k+1}-t_k)}x(t_k)+\int_{t_k}^{t_{k+1}}Be^{A(t_{k+1}-s)} u(s) ds
\end{equation}

By ZOH, we set $u(s)=u(t_k)$ in $[t_k,t_{k+1}]$. In addition, we have $\Delta=t_{k+1}-t_k$ by definition. Then, through easy computation,  gives the formulation of $\overline{A}$ and $\overline{B}$ in Mamba.
\begin{equation}
    x(t_{k+1})=exp(A\Delta)x(t_k)+(\Delta A)^{-1}(exp(\Delta A)-I)\Delta B u(t_k)
\end{equation}

\subsubsection{Hardware-aware Algorithm}
The naive recurrent computation uses $O\left(BLDN\right)$ FLOPs, while convolutional computation uses $O\left(BLD\log (L)\right)$ FLOPs. The recurrent computation model uses fewer FLOPs than convolution. However, this brings two issues: its sequential nature and a large amount of memory. The efficient parallel scanning algorithm addresses the sequentiality problem. To tackle the memory issue, the complete state $h$ is not stored practically. Instead, GPU's hierarchical memory structure is leveraged, storing the state $h$ only in certain efficient memory structures and using kernel fusion. Moving from slow HBM to fast SRAM, SSM parameters ($\Delta$, $A$, $B$, $C$) are discretized and computed recurrence in the SRAM. The output ($B$, $L$, $D$) is then stored back into the slow HBM.

\section{Discussion on the Substitutability}
\label{sec:substitute}
In this chapter, we have reviewed recent work on Mamba. Most of the research involves modifications to Mamba, adapting it to different specific tasks, and comparing it with existing Transformer architectures. For example, MambaByte \citep{wang2024mambabyte} improves subword Transformers by performing autoregression on bytes, enabling token-free adaptation of Mamba. MambaFormer \citep{park2024can} combines Mamba with attention blocks and applies it to in-context learning tasks. UVM-Net \citep{zheng2024u} utilizes a designed Bi-SSM module to capture local features and long-range dependencies, comparing it with the popular dehazing architecture Transformer.  The following are some representative works that may replace Vision Transformer, Graph Transformer, Swin Transformer/U-Net, etc.

\subsection{Vision Transformer}
Vision Transformer (ViT) \citep{dosovitskiy2020image} has achieved tremendous success in applying Transformer to the field of computer vision. It segments images into patches and flattens these patches, effectively treating them as sequence tokens similar to those in natural language processing tasks.

The recently proposed Vision Mamba (Vim) \citep{zhu2024vision} outperforms DeiT in ImageNet1k image classification tasks. It demonstrates a 2.8x improvement in processing high-resolution images compared to DeiT, with an 86.8$\%$ reduction in GPU memory consumption. The core of Vision Mamba lies in the introduction of a bidirectional SSM for global visual context modeling. Vmamba    \citep{liu2024vmamba} introduces the Cross Scan Module (CSM) to consider spatial relationships on the two-dimensional plane. Vmamba exhibits linear complexity and a global receptive field, demonstrating competitiveness in comparisons with DeiT and Swin Transformer.

Transformer employs MLP for channel mixing and attention for sequence mixing. In contrast, SiMBA \citep{patro2024simba} introduces Einstein FFT (EinFFT) for channel modeling through specific eigenvalue computations, and utilizes the Mamba block for sequence modeling. 

\begin{table}[h]
\label{table1}
\centering
	\caption{\textbf{Accuracy comparison across various models on ImageNet-1K.} The results of ResNet, ResNetXt50, RegNetY, ViT, DeiT, Swin, S4ND, Vim, VMamba and SiMBA were referenced from \citep{liu2024vmamba},\citep{zhu2024vision} and \citep{patro2024simba}.}
\begin{tabular}{ccccc}
\toprule[1.25pt]
\multicolumn{1}{c}{\textbf{method}}  & \textbf{image size}    & \textbf{\#param.} & \multicolumn{1}{c}{\textbf{FLOPs}}   & \textbf{ImageNet top-1 acc} \\ \midrule[0.75pt]
\multicolumn{5}{c}{\textbf{Convnets}}                                                                                                                    \\ \midrule[0.75pt]
\multicolumn{1}{c}{ResNet-18}        & 224 $\times$ 224 & 12M               & \multicolumn{1}{c}{1.8G}             & 69.8                        \\
\multicolumn{1}{c}{ResNet-50}        & 224 $\times$ 224 & 25M               & \multicolumn{1}{c}{3.8G}             & 76.2                        \\
\multicolumn{1}{c}{ResNet-101}       & 224 $\times$ 224 & 45M               & \multicolumn{1}{c}{7.6G}             & 77.4                        \\
\multicolumn{1}{c}{Res-Net-152}      & 224 $\times$ 224 & 60M               & \multicolumn{1}{c}{11.3G}            & 78.3                        \\ \midrule[0.5pt]
\multicolumn{1}{c}{ResNetXt50-32x4d} & 224 $\times$ 224 & 25M               & \multicolumn{1}{c}{4.2G}             & 77.6                        \\ \midrule[0.5pt]
\multicolumn{1}{c}{RegNetY-4G}       & 224 $\times$ 224 & 21M               & \multicolumn{1}{c}{4.0G}             & 80.0                        \\
\multicolumn{1}{c}{RegNetY-8G}       & 224 $\times$ 224 & 39M               & \multicolumn{1}{c}{8.0G}             & 81.7                        \\
\multicolumn{1}{c}{RegNetY-16G}      & 224 $\times$ 224 & 84M               & \multicolumn{1}{c}{16.0G}            & 82.9                        \\ \midrule[0.75pt]
\multicolumn{5}{c}{\textbf{Transformer}}                                                                                                                 \\ \midrule[0.75pt]
\multicolumn{1}{c}{ViT-B/16}         & 384 $\times$ 384 & 86M               & \multicolumn{1}{c}{55.4G}            & 77.9                        \\
\multicolumn{1}{c}{ViT-L/16}         & 384 $\times$ 384 & 307M              & \multicolumn{1}{c}{190.7G}           & 76.5                        \\ \midrule[0.5pt]
\multicolumn{1}{c}{DeiT-S}           & 224 $\times$ 224 & 22M               & \multicolumn{1}{c}{4.6G}             & 79.8                        \\
\multicolumn{1}{c}{DeiT-S}           & 224 $\times$ 224 & 86M               & \multicolumn{1}{c}{17.5G}            & 81.8                        \\
\multicolumn{1}{c}{DeIT-B}           & 384 $\times$ 384 & 86M               & \multicolumn{1}{c}{55.4G}            & 83.1                        \\ \midrule[0.5pt]
\multicolumn{1}{c}{Swin-T}           & 224 $\times$ 224 & 29M               & \multicolumn{1}{c}{4.5G}             & 81.3                        \\
\multicolumn{1}{c}{Swin-S}           & 224 $\times$ 224 & 50M               & \multicolumn{1}{c}{8.7G}             & 83.0                        \\
\multicolumn{1}{c}{Swin-B}           & 224 $\times$ 224 & 88M               & \multicolumn{1}{c}{15.4G}            & 83.5                        \\ \midrule[0.75pt]
\multicolumn{5}{c}{\textbf{SSMs}}                                                                                                                        \\ \midrule[0.75pt]
\multicolumn{1}{c}{S4ND-ViT-B}       & 224 $\times$ 224 & 89M               & \multicolumn{1}{c}{-} & 80.4                        \\ \midrule[0.5pt]
\multicolumn{1}{c}{Vim-Ti}           & 224 $\times$ 224 & 7M                & \multicolumn{1}{c}{-} & 73.1                        \\
\multicolumn{1}{c}{Vim-S}            & 224 $\times$ 224 & 26M               & \multicolumn{1}{c}{-} & 80.3                        \\ \midrule[0.5pt]
\multicolumn{1}{c}{VMamba-T}         & 224 $\times$ 224 & 22M               & \multicolumn{1}{c}{4.5G}             & 82.5                        \\
\multicolumn{1}{c}{VMamba-S}           & 224 $\times$ 224 & 44M               & \multicolumn{1}{c}{9.1G}             & 83.6                        \\
\multicolumn{1}{c}{VMamba-B}         & 224 $\times$ 224 & 75M               & \multicolumn{1}{c}{15.2G}            & 83.9              
                  \\ \midrule[0.5pt]
\multicolumn{1}{c}{SiMBA-S(MLP)}         & 224 $\times$ 224 & 26.5M               & \multicolumn{1}{c}{5.0G}             & 84.0                        \\
\multicolumn{1}{c}{SiMBA-B(MLP)}           & 224 $\times$ 224 & 40.0M               & \multicolumn{1}{c}{9.0G}             & 84.7                        \\
\multicolumn{1}{c}{SiMBA-L(EinFFT)}         & 224 $\times$ 224 & 36.6M               & \multicolumn{1}{c}{7.6G}            & 83.9                   \\ 
\bottomrule[1.25pt]
\end{tabular}
\end{table}

According to the data in Table \ref{table1}, when comparing models on the ImageNet-1K dataset, models based on Mamba have fewer parameters than transformer models of similar sizes. However, there is no clear advantage in terms of FLOPs and accuracy.

Additionally, \citeauthor{cheng2024activating} utilize Vim in the area of single image super-resolution (SISR), unleashing the potential of mamba in the context of high resolution image. Building upon Vmamba's SS2D, high-order visual state space (H-VSS) \citep{wu2024h} employs a structure similar to Transformers, where the self-attention layer is replaced by the proposed High-order 2D-selective-scan (H-SS2D). Through higher-order interactions, H-VSS gradually reduces the introduction of redundant information during the SS2D operation process. EfficientVMamba \citep{pei2024efficientvmamba} enhances the selective scanning approach of Vmamba's CSM by incorporating an atrous-based strategy for efficient skip sampling. ZigMamba \citep{hu2024zigma} considers spatial continuity within the scanning scheme, similarly improving upon Vmamba's scanning methodology. LocalMamba \citep{huang2024localmamba} also modifies CSM by partitioning tokens into distinct windows, facilitating traversal within each window.

\subsection{Swin Transformer/U-Net}
U-Net \citep{ronneberger2015u} is a widely used convolutional neural network architecture for biomedical image processing. It employs an encoder-decoder structure and uses skip connections to concatenate the corresponding feature maps from the encoder to the decoder, preserving details. Swin Transformer \citep{liu2021swin}, on the other hand, is a transformer architecture for visual images. It processes patches hierarchically, utilizing shifted windows for local self-attention, and performs well on various benchmarks.

The models obtained by transforming Mamba through a combination of U-Net and Swin Transformer are predominantly used in medical image processing.
U-Mamba \citep{ma2024u} stands out as the pioneering network to integrate the Mamba framework into biomedical image analysis tasks. It employs a hybrid CNN-SSM (Structured Self-Attention) module and exhibits superior performance on datasets for organ segmentation, instrument segmentation, and cell segmentation when compared to state-of-the-art methods based on CNN and Transformer. Building upon U-Mamba, Swin-UMamba \citep{liu2024swin} incorporates a pre-training strategy based on ImageNet. The training results on the aforementioned datasets not only surpass models based on CNN and ViT but also outperform the original U-Mamba model.
SegMamba \citep{xing2024segmamba} and nnMamba \citep{gong2024nnmamba} are both applied to 3D biomedical images. SegMamba, which enhances U-Net, demonstrates the ability to process volumetric features at a resolution of 64 × 64 × 64. It surpasses the performance of the Transformer-based approach, SwinUNETR-V2, on the BraTS2023 brain tumor segmentation dataset. On the other hand, nnMamba incorporates the Mamba block into CNNs and excels in various tasks such as 3D image segmentation, classification, and landmark detection, outperforming current state-of-the-art methods. In the domain of cross-modal registration for deformable Magnetic Resonance (MR) and Computed Tomography (CT), MambaMorph \citep{guo2024mambamorph} inherits from VoxelMorph, which is based on U-Net, and exhibits superior performance compared to the Transformer-based TransMorph on the SR-Reg dataset. Mamba-UNet \citep{wang2024mamba}, employing the VMamba architecture with injected skip connections, outperforms UNet and SwinUNet on an MRI dataset for multi-structure heart segmentation. Semi-Mamba-UNet \citep{wang2024semi} integrates a visual mamba-based UNet architecture with a conventional UNet into a semi-supervised learning (SSL) framework to generate pseudo-labels and cross-supervise each other. It is compared with various SSL frameworks on an MRI dataset. Weak-Mamba-UNet \citep{wang2024weak} is a weakly-supervised learning (WSL) framework that incorporates UNet, SwinUNet, and Mamba-UNet for efficient long-range dependency modeling, stacking structured state-space models hierarchically to address raw sensor data prediction tasks. VM-UNet \citep{ruan2024vm}, rooted in a pure SSM model, combines elements from VMamba and U-Net. Experimental results on the ISIC17, ISIC18, and Synapse datasets showcase its competitiveness in biomedical image analysis tasks. The subsequent introduction of VM-UNetV2 \citep{zhang2024vm} improved the performance of VM-UNet. 

\begin{table}[htb]
\setlength{\tabcolsep}{1pt}
\centering
{
\caption{\textbf{Results of different U-Net-based Mamba models across various datasets.} Provide only one example of transformer-based models for reference. * means using a Mamba-based decoder. The outcomes of UMamba and Swin-UMamba were cited from \citep{liu2021swin}. SegMamba's results were sourced from \citep{xing2024segmamba}. The results of MambaMorph were obtained from \citep{guo2024mambamorph}. Mamba-UNet's results were derived from \citep{wang2024mamba}. Semi-Mamba-UNet's outcomes were referenced from \citep{wang2024semi}. VM-UNet's results were cited from \citep{ruan2024vm}.}
\label{tab:agent_family}
\scalebox{0.75}{
\begin{tabular}{lccccccccccc}
\toprule[1.25pt]
\multirow{2}{*}{Datasets}                                        
& \multirow{2}{*}{Methods}          
& \multicolumn{2}{c}{Based Models}
& \multirow{2}{*}{  Transformer-based}  
& \multirow{2}{*}{Domain}             
& \multicolumn{5}{c}{Evaluation Metrics} 

\\ \cline{3-4} 
\cline{7-11} 
&                                
& U-Net                              
& Others          
&  
&
& Dice↑           
& HD95↓
& DSC↑
& NSD↑
& F1↑
\\  \midrule
\midrule
Abdomen MRI
& U-Mamba$\_$Bot                          
& $\checkmark$                  
& - 
& UNETR
& Organ segmentation
& -
& -
& 0.7588           
& 0.8285\\
\midrule
Abdomen MRI
& U-Mamba$\_$Enc                          
& $\checkmark$                  
& -
& UNETR
& Organ segmentation
& -
& -
& 0.7625           
& 0.8327
&-\\
\midrule
Abdomen MRI
& Swin-UMamba$^*$                       
& $\checkmark$                  
& U-Mamba
& SwinUNETR
& Organ segmentation
&-
&-
& 0.7705           
& 0.8376
&-\\
\midrule
Abdomen MRI
&Swin-UMamba                          
& $\checkmark$                  
& U-Mamba
& SwinUNETR
& Organ segmentation
&-
&-
& 0.7768           
& 0.8442
&-\\
\midrule
\midrule
Endoscopy
& U-Mamba$\_$Bot                          
& $\checkmark$                   
& -
& UNETR
& Instrument segmentation
& -
& -
& 0.6540           
& 0.6692
& -\\
\midrule
Endoscopy
& U-Mamba$\_$Enc                          
& $\checkmark$                   
& -
& UNETR
& Instrument segmentation
& -
& -
& 0.6303           
& 0.6451
& -\\
\midrule
Endoscopy
& Swin-UMamba$^*$                         
& $\checkmark$                  
& U-Mamba
& SwinUNETR
& Instrument segmentation
& -
& -
& 0.6717           
& 0.6871
& -\\
\midrule
Endoscopy
& Swin-UMamba                         
& $\checkmark$                  
& U-Mamba
& SwinUNETR
& Instrument segmentation
& -
& -
& 0.6774           
& 0.6926
& -\\
\midrule
\midrule
Microscpoy
&U-Mamba\_Bot                          
& $\checkmark$                  
& -
& UNETR
& Cell segmentation
& -
& -
& -           
& -
& 0.5389\\
\midrule
Microscpoy
&U-Mamba$\_$Enc                         
& $\checkmark$                  
& -
& UNETR
& Cell segmentation
& -
& -
& -           
& -
& 0.5607\\
\midrule
Microscpoy
&Swin-UMamba$^*$                        
& $\checkmark$                  
& U-Mamba
& SwinUNETR
& Cell segmentation
& -
& -
& -           
& -
& 0.5850\\
\midrule
Microscpoy
&Swin-UMamba                          
& $\checkmark$                  
& U-Mamba
& SwinUNETR
& Cell segmentation
& -
& -
& -           
& -
& 0.5836\\
\midrule
\midrule
BraTS2023 
& SegMamba                          
& $\checkmark$                         
& -
& nnFormer
& 3D biomedical images                               
& 0.9132           
& 3.5600
&-\\
\midrule
\midrule
SR-Reg 
&MambaMorph                          
& $\checkmark$                         
& VoxelMorph 
& TransMorph
& Deformable image registration                       
& 0.8271           
& 2.0000
& -
& -
&-\\
\midrule
\midrule
MRI Cardic Test Set 
&Mamba-UNet                          
& $\checkmark$
& VMamba     
& Swin-UNet
& 2D biomedical images                               
& 0.9281           
& 2.4645    
& -
& -
&-\\
\midrule
MRI Cardic Test Set
&Semi-Mamba-UNet                          
& $\checkmark$                       
& SSL framework 
& -
& Medical image segmentation                    
& 0.8396           
& 6.2139
& -
& -
&-\\
\midrule
MRI Cardic Test Set
&Weak-Mamba-UNet                           
& $\checkmark$                      
& WSL framework   
& -
& Medical image segmentation                               
& 0.9171           
& 3.9597
& -
& -
&-\\
\midrule
\midrule
ISIC17
&VM-UNet
& $\checkmark$                     
& VMamba  
& Swin-UNet
& 2D medical image segmentation                          
& -           
& -
& 89.03
& -
&-\\
\midrule
ISIC18
&VM-UNet
& $\checkmark$                       
& VMamba   
& Swin-UNet
& 2D medical image segmentation                          
& -           
& -
& 89.71
& -
&-\\
\midrule
Synapse
&VM-UNet
& $\checkmark$                      
& VMamba    
& Swin-UNet
& 2D medical image segmentation                          
& -           
& 19.21
& 81.08
& -
&-
\\\bottomrule[1.25pt]
\end{tabular}
}
}
\end{table}

\subsection{Graph Transformer}
Graph Neural Networks (GNNs) suffer from drawbacks such as excessive compression and inadequate capture of long-range dependencies. Currently, the most widely used is Graph Transformer (GT) \citep{yun2019graph}. However, the high computational cost of GT makes it perform well on small-scale datasets but limits its applicability to larger datasets. Two proposed solutions are sparse attention and subgraph tokenization. The GraphGPS framework \citep{rampavsek2022recipe} allows the replacement of fully connected with sparse attention, resulting in $O(N + E)$ complexity. However, this requires costly Laplacian eigen-decomposition (PE) and structural encoding (SE) to achieve optimal performance, still bringing the complexity to $O(N^2)$ \citep{behrouz2024graphmamba}. On the other hand, passing subgraph tokens to Message-Passing Neural Networks (MPNNs) \citep{gilmer2020message} may lead to issues such as over-smoothing \citep{rusch2023survey}, over-squashing \citep{di2023over}, and poor capture of long-range dependencies \citep{dwivedi2022long}. To address these challenges, Mamba is introduced as a solution with linear complexity.

Graph-Mamba \citep{wang2024graph-mamba} adopts the GPS framework and incorporates a designed Graph-Mamba block (GMB). The GMB employs a heuristic sorting of nodes based on the idea that more crucial nodes should have access to more context. This heuristic sorting adjusts the priority of input nodes, allowing for improved processing of important nodes with more context. The results demonstrate that Graph-Mamba outperforms in long-range graph prediction tasks.

Graph Mamba \citep{behrouz2024graphmamba} introduces a graph tokenization process that produces a long sequence of diverse subgraphs. Let $G=(V, E)$ represent a graph, where $V={v_{1}, \ldots, v_{n}}$ denotes the set of nodes and $E \subseteq V \times V$ is the set of edges. We assume each node $ v \in V$ has a subset of nodes $S \subseteq V$ as the set of the neighbors of $v$. We employ $G[S]$ to signify the induced subgraph formed by nodes in set $S$. Starting from node $v$, randomly sample paths of length $\hat{m}$. Let $T_{\hat{m}, i}(v) (i=0,\dots, M)$ denote the set of all nodes on this sampled path. In equation \ref{subgraph}, $G\left[T_{\hat{m}}(v)\right]$ represents the subgraph of a sample of the $\hat{m}$-hop neighborhood of node $v$ and the sequence of $G\left[T_{\hat{m}}(v)\right]$  corresponds to the tokens of node $v$.
\begin{equation}
    \label{subgraph}
    G\left[T_{\hat{m}}(v)\right]=G\left[\bigcup_{i=1}^{M} T_{\hat{m}, i}(v)\right]
\end{equation}
Subsequently, employ an encoder $\phi$ to encode the subgraph of node $v$ in equation \ref{encode}, where ${x}_v^{0} \in \mathbf{X}$ is the feature vector corresponding to each node, and $\mathbf{X} \in \mathbb{R}^{n \times d}$ is the feature matrix describing the nodes.
\begin{equation}
    \label{encode}
   x_v^{((i-1)s+j)}=\phi\left(G\left[T_{i}^{j}(v)\right],\mathbf{X}_{T_{i}^{j}(v)}\right)
\end{equation}

Graph Mamba also proposes bidirectional State Space Models (SSMs) for graphs, utilizing forward and backward scans, which aligns with the approach proposed in earlier works \citep{zhu2024vision, wang2022pretraining}. This bidirectional approach ensures that preceding tokens contain information from succeeding tokens and vice versa. Graph Mamba achieves outstanding performance on long-range, small-scale, large-scale, and anisotropic benchmark datasets while consuming less GPU memory.

\subsection{Others}
The remaining improvements for Mamba primarily focus on three aspects. Firstly, enhancements are made to the Mamba architecture itself. Most works involve integrating different modules to combine the ability of SSM to capture long-range dependencies with other modules to complete downstream tasks. For example, FD-Vision Mamba \citep{zheng2024fd} integrates the advantages of CNN local pattern recognition and SSM for endoscope exposure correction tasks. Mamba models combined with MoE  \citep{pioro2024moe, anthony2024blackmamba} address the issue of high memory usage in MoE. LOCOST \citep{bronnec2024locost} combines state space models with an encoder-decoder architecture, achieving a computational complexity of $O(L\log L)$, which outperforms models based on sparse attention patterns and enables the handling of longer input sequences. DenseMamba \citep{he2024densemamba} selectively integrates shallow hidden states into deeper layers, preserving fine-grained information in the final output.

Secondly, there are modifications to handle different types of data objects that Mamba processes.
MambaTab \citep{ahamed2024mambatab} is designed to handle tabular data and demonstrates superior performance compared to state-of-the-art baselines, including Transformers and CNN-based models. Mamba is a suitable  for managing intricate long structures, such as genomics \citep{schiff2024caduceus}, audio\citep{quan2024multichannel}, and video \citep{chen2024video,li2024videomamba}. 
Recently, there have been proposals to extend one-dimensional Mamba models to higher dimensions. For instance, Point Mamba \citep{liang2024pointmamba}  enhances Point Cloud Transformers by proposing a reordering strategy that strengthens the global modeling capability of SSM through geometric scanning order. The model proposed by \citeauthor{liu2024point} also deals with point clouds and devises a sorting strategy based on octrees for the raw irregular points. 
STG-Mamba \citep{li2024stg} is designed for processing dynamic Spatial-Temporal Graph (STG) data. Additionally, Mamba-ND \citep{li2024mamba} extends the Mamba architecture to handle arbitrary multi-dimensional data.

Thirdly, Mamba-based models are employed across various domains.

\begin{table}[h]
\label{table3}
\centering
	\caption{\textbf {A summary of the various fields where Mamba-based models are applied.}}
\begin{tabular}{ll}
\toprule[1.25pt]
\multicolumn{1}{c}{\textbf{domain}}  & \multicolumn{1}{c}{\textbf{model}}     

\\ \midrule[0.75pt] 
\multicolumn{2}{c}{\textbf{{Vision}}}              \\ \midrule[0.75pt]

\multicolumn{1}{l}{Image generation}  &\citep{hu2024zigma, zheng2024fd, zhang2024motion}\\ \multicolumn{1}{c}{} &
\citep{zhu2024dig, fei2024dimba, teng2024dim, mo2024scaling}\\
\multicolumn{1}{l}{Image fusion}  &\citep{cao2024novel, peng2024fusionmamba}\\
\multicolumn{1}{l}{Image restoration}  &\citep{guo2024mambair, fu2024hdmba, zheng2024u, li2024fouriermamba} \\
\multicolumn{1}{l}{}  
&\citep{zhen2024freqmamba, zhou2024rsdehamba, deng2024cumamba} \\
\multicolumn{1}{l}{Image enhancement}  &\citep{lin2024pixmamba, zhang2024llemamba, bai2024retinexmamba}\\
\multicolumn{1}{l}{Image classification}  &\citep{chen2024res, sheng2024dualmamba, zhou2024mambainmamba, yao2024spectralmamba} \\
\multicolumn{1}{l}{}  
&\citep{chen2024rsmamba, yue2024medmamba} \\
\multicolumn{1}{l}{Image segmentation}  &\citep{khan2024convolution, tsai2024uumamba, yuan2024mucmnet, liu2024cmunet} \\
\multicolumn{1}{c}{} &
\citep{archit2024vimunet, ma2024rs3mamba, yang2024remamber, tang2024rotate}\\
\multicolumn{1}{c}{} &
\citep{wu2024h, xie2024promamba, zhang2024vm, liao2024lightmunet}\\
\multicolumn{1}{c}{} &
\citep{ma2024u, ruan2024vm}\\
\multicolumn{1}{l}{Object detection}  &\citep{wang2024mambayolo, chen2024mim, ma2024feryolomamba, verma2024soar} \\
\multicolumn{1}{l}{Change detection}  &\citep{zhang2024cdmamba, chen2024changemamba}\\
\multicolumn{1}{l}{Video processing}  &\citep{yang2024vivim, gao2024matten, chaudhuri2024simba}\\
\multicolumn{1}{c}{} &
\citep{chen2024video, li2024videomamba, chen2024demamba}\\

\midrule[0.75pt]
\multicolumn{2}{c}{\textbf{{Language}}}
\\ \midrule[0.75pt]

\multicolumn{1}{l}{LLMs}  &\citep{yang2024clinicalmamba, lieber2024jamba, ren2024samba, he2024densemamba} \\
\multicolumn{1}{l}{} &\citep{de2024griffin}\\

\midrule[0.75pt]
\multicolumn{2}{c}{\textbf{{Mutli-modal}}}
 \\ \midrule[0.75pt]

\multicolumn{1}{l}{Multi-modal LLMs}  &\citep{qiao2024vl, zhao2024cobra, liu2024robomamba} \\
\multicolumn{1}{l}{Multi-modality object detection}  &\citep{dong2024fusionmamba} \\
\multicolumn{1}{l}{Multi-modality image fusion}  &\citep{li2024mambadfuse} \\
\multicolumn{1}{l}{Multi-modality semantic segmentation}  &\citep{wan2024sigma} \\
\multicolumn{1}{l}{Multi-modality 3D medical images}  &\citep{yang2024cmvim} \\
\multicolumn{1}{l}{Image data \& genomic data}  &\citep{zhou2024mgi, chen2024survmamba} \\
\multicolumn{1}{l}{RGB-Event based tracking}  &\citep{huang2024mambafetrack} \\

\midrule[0.75pt]
\multicolumn{2}{c}{\textbf{{Others}}}
\\ \midrule[0.75pt]

\multicolumn{1}{l}{Time series forecasting}  &\citep{wang2024Time, ahamed2024timemachine, xu2024integrating, shi2024mambastock} \\
\multicolumn{1}{l}{} &\citep{zeng2024cmamba, liang2024bimamba, wang2024Time}\\
\multicolumn{1}{l}{Point cloud}
&\citep{liang2024pointmamba, liu2024point, chen2024pointabmintegrating,han2024mamba3d}\\
\multicolumn{1}{l}{} &\citep{wang2024pointramba, liu2024mamba4d, li20243dmambacomplete, zhou20243dmambaipf}\\
\multicolumn{1}{l}{} &\citep{zhang2024point}\\
\multicolumn{1}{l}{Sequential recommendation}  &\citep{liu2024mamba4rec, wang2024echomamba4rec} \\
\multicolumn{1}{l}{Audio processing}  &\citep{chen2024rawbmamba, erol2024audio, lin2024audio, shams2024ssamba} \\
\multicolumn{1}{l}{Speech processing}  
&\citep{zhang2024mambaspeech, chao2024investigation, li2024spmamba, jiang2024dualpath} \\
\multicolumn{1}{l}{3D processing}
&\citep{mo2024efficient, shen2024gamba} \\
\multicolumn{1}{l}{Gesture synthesis}  &\citep{xu2024mambatalk} \\
\multicolumn{1}{l}{DNA sequence modeling}  &\citep{schiff2024caduceus} \\
\multicolumn{1}{l}{4D light fields}  &\citep{xia2024lfmamba} \\
\multicolumn{1}{l}{Network traffic}  &\citep{chu2024feasibility, wang2024netmamba} \\
\multicolumn{1}{l}{Motion style transfer}  
&\citep{qian2024smcd} \\

\bottomrule[1.25pt]
\end{tabular}
\end{table}

\citeauthor{he2024pan} introduce Pan Mamba in the pan-sharpening domain, introducing Channel Swapping Mamba Block and Cross-modality Mamba Block. Res-Vmamba \citep{chen2024res} integrates a residual learning framework into the VMamba model for food classification. ClinicalMamba \citep{yang2024clinicalmamba} is a generative clinical language model pretrained on clinical notes, while Cobra \citep{zhao2024cobra} and VL-Mamba \citep{qiao2024vl} integrates the efficient Mamba language model into the visual modality.
Additionally, Mamba4Rec \citep{liu2024mamba4rec} applies Mamba in the domain of efficient sequential recommendation. In the domain of human-computer interaction, MambaTalk \citep{xu2024mambatalk} enhances gesture diversity and rhythm through multimodal integration, whereas MambaIR \citep{guo2024mambair} is utilized in the field of image restoration. Moreover, Mamba-in-Mamba \citep{chen2024mim} explores both global and local information, employed for infrared small target detection. Many models, including those proposed by  \citeauthor{hu2024zigma, zheng2024fd, zhang2024motion, fei2024dimba, teng2024dim}, integrate Mamba into diffusion models. Furthermore, Mamba is employed in time series forecasting \citep{wang2024Time, ahamed2024timemachine, xu2024integrating}, image fusion \citep{cao2024novel, peng2024fusionmamba}, and remote sensing change detection tasks \citep{zhang2024cdmamba, chen2024changemamba}. 

\subsection{Discussion}
Mamba-based models outperform Transformer-based models primarily because the Mamba architecture has linear complexity, while the attention mechanism has a complexity of $O(N^2)$. With hardware-aware algorithm, Mamba-based models have lower FLOPs. This advantage is particularly evident in high-resolution image processing, which is crucial in biomedical image processing. Moreover, when dealing with complex data such as large graphs and multi-dimensional data, Mamba-based models theoretically perform better.

However, research indicates that Mamba maintain a fixed-size recurrent state, but struggle at recall \citep{arora2024simple}. Mamba's potential in handling visual images has also been questioned \citep{yu2024mambaout}. Researchers conducted experiments on the ImageNet-1K classification task \citep{imagenet15russakovsky} using a model based on gated CNN blocks but without SSM, and the model outperformed architectures like VMamba.

\begin{figure*}[htb]
    \centering
    \vspace{-3mm}
complete    \includegraphics[width=1.0\textwidth]{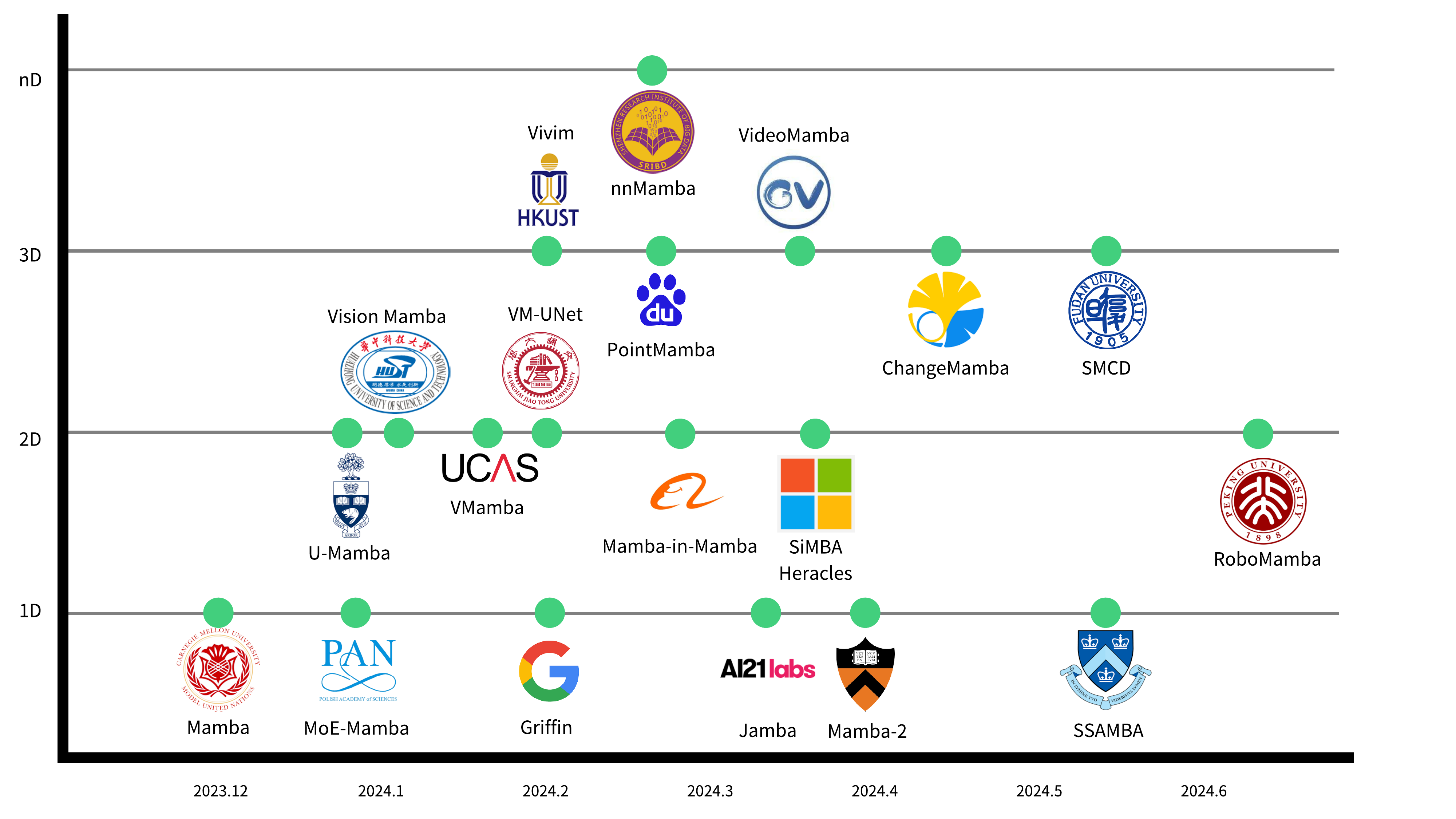}
    \vspace{-6mm}
    \caption{Representative works from the emergence of Mamba to the present \citep{gu2023mamba, pioro2024moe, de2024griffin, lieber2024jamba, dao2024transsm, shams2024ssamba, ma2024u, zhu2024vision, liu2024vmamba, ruan2024vm, chen2024mim, patro2024simba, Heracles, liu2024robomamba, yang2024vivim, liang2024pointmamba, li2024videomamba, chen2024changemamba, qian2024smcd, gong2024nnmamba}.}
    \label{represent}
    \vspace{-1mm}
\end{figure*}

\section{The Combination of Transformer and Mamba}
\label{combine}
Despite Mamba's impressive performance in long sequences and autoregressive tasks, it still has certain issues. It requires sophisticated initialization techniques and specialized implementations to achieve high quality and runtime performance \citep{fu2023simple}, which lead to stability problems when scaling SSM to large network sizes. Also, Mamba is more efficient to train than RNNs but how to train SSMs as efficiently as Transformers remains a challenge. However, recent studies suggest that there are now theoretical explanations bridging the conceptual gap between SSMs and attention-based models \citep{dao2024transsm}. Moreover, an increasing number of models combining Transformers and SSMs are being proposed, indicating that this integration might become a major trend in the future.

\subsection{Transformers are SSMs}
\subsubsection{Softmax Self-attention}
The most common variant of attention is softmax attention. Given query Q, key K, and value V matrices, the attention output is calculated as follows where $Q, K, V \in \mathbb{R}^{(T,P)}$:

\begin{equation}
    Y=softmax(QK^T)\cdot V
\end{equation}
where T denotes the sequence or time axis, with subscripts indexing into the first dimension, for example, $X_{t}$, $Y_{t}$
 $\in \mathbb{R}^{P}$.

\subsubsection{SSD Framework} The state space dual (SSD) is made through the abstractions of structured matrices \citep{dao2024transsm}. SSD layer can be considered a specific form of the selective SSM . The diagonal structured A used in Mamba is further simplified to a scalar multiplied by an identity structure. $a_{i}$ are input-dependent scalars bounded in [0, 1] and $L$ is an additional mask matrix. The dual form can be represented as:
\begin{equation}
    (L\circ QK^T)\cdot V, \; L_{ij}= \left\{
    \begin{array}{ll}
        a_{i} \times \dots \times a_{j+1}  &i\geq j \\ 
        0 & i < j \\
    \end{array}
    \right.
\end{equation}\label{ssd}

$L$ can be viewed as replacing the heuristic positional embeddings of the Transformer with masks related to different data positions. $a_{i}$ can also be viewed as the selective mechanism in Mamba. In addition, after discarding softmax, sequence computation can be achieved using linear scaling through the linear attention framework \citep{katharopoulos2020transrnn}.

Through this mathematical explanation, researches successfully integrated the Transformer and Mamba architectures, and trained the Mamba-2 model \citep{dao2024transsm}, which addresses issues such as training instability in Mamba.

\subsection{Another Explanation: Formulate Models as Kernel Methods}
\label{math}
In this section, we will explore the recently proposed views on framing Transformer and RNN as kernel methods, which sheds light on a novel and illuminating way to analyze the performance of these networks and may lead to the birth of new update models. We will further explain that under the kernel method, Mamba will be unified with the RNN structure.
\subsubsection{A Brief Review on Kernel Method}
The tasks in real life are often not linearly separable, and a feasible method is to project the data into high dimensional feature space. However, the mapping is always not accessible or computationally expensive. Kernel method is introduced to solve this issue by directly defining the inner product in the high dimensional space without specifying the high-dimensional mapping.

The minimization of a optimization problem can utilize the kernel method thanks to the following Representer Theorem
\begin{equation}
    \mathop{\min}_{\omega}\frac{1}{n}\sum_{i=1}^n L(\omega^T \phi(x_i),y_i)+\frac{\lambda}{2}\|\omega\|_2^2, \lambda\geq 0
\end{equation}
If the above optimization has optimal solutions, then it must be of the form
\begin{equation}
    \omega^\star=\sum_{i=1}^n \alpha_i \phi(x_i)
\end{equation}

Common used kernel functions including linear kernel, polynomial kernel, Sigmoid kernel, etc. Introducing kernel methods into the analysis of deep learning has been a popular direction and more work should be done to investigate the correlations between the two methods.
\subsubsection{Transformer-type}
\citeauthor{tsai2019transformer} frames the attention in transformer as a kernel method noticing that both of them can measure the similarities of two different elements of input sequences. Specifically, the kernel-based formulation can be presented as:
\begin{equation}
    \textsc{Attention}(x_q; M(x_q,S_{x_k}))=\sum_{x_k\in M(x_q,S_{x_k})}\frac{k(x_q,x_k)}{\sum_{x_k'\in M(x_q,S_{x_k})}k(x_q,x_k)}v(x_k)
\end{equation}
where $x_q, x_k$ denote queries and keys, $v(\cdot)$ is the value function, $k(\cdot,\cdot)$ represents the given kernel function, and $S_{x_k}$ is the set representation of the keys. $M(\cdot,\cdot)$ is named as the filtering function,  decides how keys operate with queries corresponding to different Transformer models. 

A promising research field is to investigate the construction of the kernel function. A typical work is the linear attention \citep{katharopoulos2020transformers}, which replaces the softmax with the linear dot-product of a kernel feature map, the exponential linear unit $elu(\cdot)$:
\begin{equation}
    \phi(x)=elu(x) + 1
\end{equation}
In this way, the complexity is reduced from quadratic to $\mathcal{O}(N)$.
\subsubsection{RNN-type}
\citeauthor{fermanian2021framing} investigates the interpretation of RNN within the framework of continuous neural network equations, framing it as kernel methods in a suitable Reproducing Kernel Hilbert Space (RKHS). Specifically, most recurrent-structured network, including RNN, LSTM and GRU can presented as a unified framwork
\begin{equation}
    h_{j+1}=h_j+\frac{1}{T}f(h_j,x_{j+1}), 0\leq j\leq T-1
\end{equation}
where $f(\cdot)$ is the feedforward model, $h_j$ represents the hidden state and $x_j$ represents the input data. 
[todo]
This equation can be regarded as a discrete counterpart of a continuous-time ODE
\begin{equation}
    dH_t=f(H_t,X_t)dt
\end{equation}
and the approximation can be guaranteed by the proposition assuming $f$ is Lipschitz continuous in $h$ and $x$
\begin{equation}
    \|H_{j/T}-h_j\|\leq c_1/T
\end{equation}
Then rewrite the above ODE into CDE under the same assumption
\begin{equation}
    d\overline{H}_t=F(\overline{H}_t)d\overline{X}_t
\end{equation}
The step-N Taylor expansion of the solution $H$ of the above equation is presented as
\begin{equation}
    H_t^N=H_0+\sum_{k=1}^N\frac{1}{k!}\sum_{1\leq i_1,...,i_k\leq d}S_{[0,t]}^{(i_1,...,i_k)(X)}F^{i_1}\star...\star F^{i_k}(H_0)
\end{equation}
where $S_{[0,t]}^{(i_1,...,i_k)(X)}$ is the signature and $\star$ is the product as below
\begin{equation}
    F\star G=\sum_j \frac{\partial G}{\partial h_j}F_j
\end{equation}
Simplifying the signature as $S(X)$, the signature kernel in the Hilbert space $\mathcal{H}$ can be introduced
\begin{equation}
    (X,Y)\mapsto \left\langle S(\overline{X}),S(\overline{Y})\right\rangle
\end{equation}
\subsubsection{Comparison}
The kernel construction of RNN-type models and Transformer-type modes adopt different approaches. Due to the attention computation of Transformer, it is easily to be turn it into a kernel function. However, RNN needs a sophisticated analysis from discrete to continuous and from ODE to CDE and finally the signature properties in the Hilbert space.

The common idea is that modifying the formulation of the kernel is possible to be a future study field. In addition, expressing the original network architecture equations in kernel structure may aid in analyzing their stability and generalization properties.

\subsection{Transformers and SSMs: Future Trends}
There have been various efforts to combine the advantages of Transformers with those of SSMs. 
\citeauthor{zuo2022efficient} integrate an S4 layer into the base layer of the State Space Augmented Transformer and apply efficient local attention techniques to the remaining layers. \citeauthor{fathi2023block} propose a hybrid layer named the Block-State Transformer (BST), which internally combines an SSM sublayer for long-range context and a Block Transformer sublayer for short-term sequence representation. 
Jamba \citep{lieber2024jamba}, a new language model based on a hybrid Transformer-Mamba MoE architecture,  provides flexibility for balancing performance and memory requirements while maintaining high throughput. 
Subsequent developments have applied the integration of Transformer and Mamba to fields such as time series forecasting \citep{xu2024integrating}, high-resolution image processing \citep{Heracles}, text-to-image diffusion models \citep{fei2024dimba} and point cloud analysis \citep{chen2024pointabmintegrating, wang2024pointramba}. Combining the advantages of efficient training with Transformers and Mamba's strengths in handling long sequence data may become a major development direction in the future.

\section{Conclusion}
\label{conclude}
In this paper, we make a survey about Mamba including the developmental trajectory of SSM, discussion on the substitutability and discussion on the combination of Transformer and Mamba. Considering its superior linear complexity compared to the $O(N^2)$ complexity of the attention mechanism, especially in domains requiring long contexts such as high-resolution images, video, and complex multidimensional data like large graphs, Mamba has the potential to replace Transformer. However, considering the drawbacks of Mamba such as unstable training and struggles with recall, the viewpoint asserting that Mamba can replace Transformer still lacks sufficient research and experiments to conclusively support this claim. On the other hand, the combination of Transformer and Mamba may become a future development direction.




\bibliography{custom}

\begin{thebibliography}{165}
\providecommand{\natexlab}[1]{#1}
\providecommand{\url}[1]{\texttt{#1}}
\expandafter\ifx\csname urlstyle\endcsname\relax
  \providecommand{\doi}[1]{doi: #1}\else
  \providecommand{\doi}{doi: \begingroup \urlstyle{rm}\Url}\fi

\bibitem[Ahamed \& Cheng(2024{\natexlab{a}})Ahamed and
  Cheng]{ahamed2024mambatab}
Md~Atik Ahamed and Qiang Cheng.
\newblock Mambatab: A simple yet effective approach for handling tabular data.
\newblock \emph{arXiv preprint arXiv:2401.08867}, 2024{\natexlab{a}}.

\bibitem[Ahamed \& Cheng(2024{\natexlab{b}})Ahamed and
  Cheng]{ahamed2024timemachine}
Md~Atik Ahamed and Qiang Cheng.
\newblock Timemachine: A time series is worth 4 mambas for long-term
  forecasting.
\newblock \emph{arXiv preprint arXiv:2403.09898}, 2024{\natexlab{b}}.

\bibitem[Anthony et~al.(2024)Anthony, Tokpanov, Glorioso, and
  Millidge]{anthony2024blackmamba}
Quentin Anthony, Yury Tokpanov, Paolo Glorioso, and Beren Millidge.
\newblock Blackmamba: Mixture of experts for state-space models.
\newblock \emph{arXiv preprint arXiv:2402.01771}, 2024.

\bibitem[Archit \& Pape(2024)Archit and Pape]{archit2024vimunet}
Anwai Archit and Constantin Pape.
\newblock Vim-unet: Vision mamba for biomedical segmentation, 2024.

\bibitem[Arora et~al.(2024)Arora, Eyuboglu, Zhang, Timalsina, Alberti, Zinsley,
  Zou, Rudra, and R{\'e}]{arora2024simple}
Simran Arora, Sabri Eyuboglu, Michael Zhang, Aman Timalsina, Silas Alberti,
  Dylan Zinsley, James Zou, Atri Rudra, and Christopher R{\'e}.
\newblock Simple linear attention language models balance the recall-throughput
  tradeoff.
\newblock \emph{arXiv preprint arXiv:2402.18668}, 2024.

\bibitem[Bai et~al.(2024)Bai, Yin, He, Li, and Zhang]{bai2024retinexmamba}
Jiesong Bai, Yuhao Yin, Qiyuan He, Yuanxian Li, and Xiaofeng Zhang.
\newblock Retinexmamba: Retinex-based mamba for low-light image enhancement,
  2024.

\bibitem[Behrouz \& Hashemi(2024)Behrouz and Hashemi]{behrouz2024graphmamba}
Ali Behrouz and Farnoosh Hashemi.
\newblock Graph mamba: Towards learning on graphs with state space models.
\newblock \emph{arXiv preprint arXiv:2402.08678}, 2024.

\bibitem[Brogan(1991)]{brogan1991modern}
William~L Brogan.
\newblock \emph{Modern control theory}.
\newblock Pearson education india, 1991.

\bibitem[Bronnec et~al.(2024)Bronnec, Duong, Ravaut, Allauzen, Chen, Guigue,
  Lumbreras, Soulier, and Gallinari]{bronnec2024locost}
Florian~Le Bronnec, Song Duong, Mathieu Ravaut, Alexandre Allauzen, Nancy~F
  Chen, Vincent Guigue, Alberto Lumbreras, Laure Soulier, and Patrick
  Gallinari.
\newblock Locost: State-space models for long document abstractive
  summarization.
\newblock \emph{arXiv preprint arXiv:2401.17919}, 2024.

\bibitem[Cao et~al.(2024)Cao, Wu, Deng, and Zhong]{cao2024novel}
Zihan Cao, Xiao Wu, Liang-Jian Deng, and Yu~Zhong.
\newblock A novel state space model with local enhancement and state sharing
  for image fusion.
\newblock \emph{arXiv preprint arXiv:2404.09293}, 2024.

\bibitem[Chao et~al.(2024)Chao, Cheng, Quatra, Siniscalchi, Yang, Fu, and
  Tsao]{chao2024investigation}
Rong Chao, Wen-Huang Cheng, Moreno~La Quatra, Sabato~Marco Siniscalchi,
  Chao-Han~Huck Yang, Szu-Wei Fu, and Yu~Tsao.
\newblock An investigation of incorporating mamba for speech enhancement, 2024.

\bibitem[Chaudhuri \& Bhattacharya(2024)Chaudhuri and
  Bhattacharya]{chaudhuri2024simba}
Soumyabrata Chaudhuri and Saumik Bhattacharya.
\newblock Simba: Mamba augmented u-shiftgcn for skeletal action recognition in
  videos, 2024.

\bibitem[Chen et~al.(2024{\natexlab{a}})Chen, Chen, Zhou, Jiang, and
  Chen]{chen2024res}
Chi-Sheng Chen, Guan-Ying Chen, Dong Zhou, Di~Jiang, and Dai-Shi Chen.
\newblock Res-vmamba: Fine-grained food category visual classification using
  selective state space models with deep residual learning.
\newblock \emph{arXiv preprint arXiv:2402.15761}, 2024{\natexlab{a}}.

\bibitem[Chen et~al.(2024{\natexlab{b}})Chen, Huang, Xu, Pei, Chen, Li, Wang,
  Li, Lu, and Wang]{chen2024video}
Guo Chen, Yifei Huang, Jilan Xu, Baoqi Pei, Zhe Chen, Zhiqi Li, Jiahao Wang,
  Kunchang Li, Tong Lu, and Limin Wang.
\newblock Video mamba suite: State space model as a versatile alternative for
  video understanding.
\newblock \emph{arXiv preprint arXiv:2403.09626}, 2024{\natexlab{b}}.

\bibitem[Chen et~al.(2024{\natexlab{c}})Chen, Hong, Huang, Xu, Gu, Li, Lan,
  Zhu, Zhang, Wang, and Li]{chen2024demamba}
Haoxing Chen, Yan Hong, Zizheng Huang, Zhuoer Xu, Zhangxuan Gu, Yaohui Li, Jun
  Lan, Huijia Zhu, Jianfu Zhang, Weiqiang Wang, and Huaxiong Li.
\newblock Demamba: Ai-generated video detection on million-scale genvideo
  benchmark, 2024{\natexlab{c}}.

\bibitem[Chen et~al.(2024{\natexlab{d}})Chen, Song, Han, Xia, and
  Yokoya]{chen2024changemamba}
Hongruixuan Chen, Jian Song, Chengxi Han, Junshi Xia, and Naoto Yokoya.
\newblock Changemamba: Remote sensing change detection with spatio-temporal
  state space model.
\newblock \emph{arXiv preprint arXiv:2404.03425}, 2024{\natexlab{d}}.

\bibitem[Chen et~al.(2024{\natexlab{e}})Chen, Chen, Liu, Li, Zou, and
  Shi]{chen2024rsmamba}
Keyan Chen, Bowen Chen, Chenyang Liu, Wenyuan Li, Zhengxia Zou, and Zhenwei
  Shi.
\newblock Rsmamba: Remote sensing image classification with state space model,
  2024{\natexlab{e}}.

\bibitem[Chen et~al.(2024{\natexlab{f}})Chen, Tan, Gong, Chu, Wu, Liu, Ye, and
  Yu]{chen2024mim}
Tianxiang Chen, Zhentao Tan, Tao Gong, Qi~Chu, Yue Wu, Bin Liu, Jieping Ye, and
  Nenghai Yu.
\newblock Mim-istd: Mamba-in-mamba for efficient infrared small target
  detection.
\newblock \emph{arXiv preprint arXiv:2403.02148}, 2024{\natexlab{f}}.

\bibitem[Chen et~al.(2024{\natexlab{g}})Chen, Xie, Lin, Song, Yang, and
  Yu]{chen2024survmamba}
Ying Chen, Jiajing Xie, Yuxiang Lin, Yuhang Song, Wenxian Yang, and Rongshan
  Yu.
\newblock Survmamba: State space model with multi-grained multi-modal
  interaction for survival prediction, 2024{\natexlab{g}}.

\bibitem[Chen et~al.(2024{\natexlab{h}})Chen, Yi, Xue, Wang, Zhang, Dong, Zeng,
  Tao, Zhao, and Fan]{chen2024rawbmamba}
Yujie Chen, Jiangyan Yi, Jun Xue, Chenglong Wang, Xiaohui Zhang, Shunbo Dong,
  Siding Zeng, Jianhua Tao, Lv~Zhao, and Cunhang Fan.
\newblock Rawbmamba: End-to-end bidirectional state space model for audio
  deepfake detection, 2024{\natexlab{h}}.

\bibitem[Cheng et~al.(2024)Cheng, Wang, and Sun]{cheng2024activating}
Cheng Cheng, Hang Wang, and Hongbin Sun.
\newblock Activating wider areas in image super-resolution.
\newblock \emph{arXiv preprint arXiv:2403.08330}, 2024.

\bibitem[Chu et~al.(2024)Chu, Jiang, Liu, Bhagoji, Bronzino, Schmitt, and
  Feamster]{chu2024feasibility}
Andrew Chu, Xi~Jiang, Shinan Liu, Arjun Bhagoji, Francesco Bronzino, Paul
  Schmitt, and Nick Feamster.
\newblock Feasibility of state space models for network traffic generation,
  2024.

\bibitem[Dao(2023)]{dao2023flashattention}
Tri Dao.
\newblock Flashattention-2: Faster attention with better parallelism and work
  partitioning.
\newblock \emph{arXiv preprint arXiv:2307.08691}, 2023.

\bibitem[Dao \& Gu(2024)Dao and Gu]{dao2024transsm}
Tri Dao and Albert Gu.
\newblock Transformers are ssms: Generalized models and efficient algorithms
  through structured state space duality.
\newblock \emph{arXiv preprint arXiv:2405.21060}, 2024.

\bibitem[Dao et~al.(2022)Dao, Fu, Ermon, Rudra, and
  R{\'e}]{dao2022flashattention}
Tri Dao, Dan Fu, Stefano Ermon, Atri Rudra, and Christopher R{\'e}.
\newblock Flashattention: Fast and memory-efficient exact attention with
  io-awareness.
\newblock \emph{Advances in Neural Information Processing Systems},
  35:\penalty0 16344--16359, 2022.

\bibitem[De et~al.(2024)De, Smith, Fernando, Botev, Cristian-Muraru, Gu,
  Haroun, Berrada, Chen, Srinivasan, et~al.]{de2024griffin}
Soham De, Samuel~L Smith, Anushan Fernando, Aleksandar Botev, George
  Cristian-Muraru, Albert Gu, Ruba Haroun, Leonard Berrada, Yutian Chen,
  Srivatsan Srinivasan, et~al.
\newblock Griffin: Mixing gated linear recurrences with local attention for
  efficient language models.
\newblock \emph{arXiv preprint arXiv:2402.19427}, 2024.

\bibitem[Dean \& Kanazawa(1989)Dean and Kanazawa]{dean1989model}
Thomas Dean and Keiji Kanazawa.
\newblock A model for reasoning about persistence and causation.
\newblock \emph{Computational intelligence}, 5\penalty0 (2):\penalty0 142--150,
  1989.

\bibitem[Deng \& Gu(2024)Deng and Gu]{deng2024cumamba}
Rui Deng and Tianpei Gu.
\newblock Cu-mamba: Selective state space models with channel learning for
  image restoration, 2024.

\bibitem[Devlin et~al.(2018)Devlin, Chang, Lee, and Toutanova]{devlin2018bert}
Jacob Devlin, Ming-Wei Chang, Kenton Lee, and Kristina Toutanova.
\newblock Bert: Pre-training of deep bidirectional transformers for language
  understanding.
\newblock \emph{arXiv preprint arXiv:1810.04805}, 2018.

\bibitem[Di~Giovanni et~al.(2023)Di~Giovanni, Giusti, Barbero, Luise, Lio, and
  Bronstein]{di2023over}
Francesco Di~Giovanni, Lorenzo Giusti, Federico Barbero, Giulia Luise, Pietro
  Lio, and Michael~M Bronstein.
\newblock On over-squashing in message passing neural networks: The impact of
  width, depth, and topology.
\newblock In \emph{International Conference on Machine Learning}, pp.\
  7865--7885. PMLR, 2023.

\bibitem[Dong et~al.(2024)Dong, Zhu, Lin, Luo, Shen, Liu, Zhang, Guo, and
  Zhang]{dong2024fusionmamba}
Wenhao Dong, Haodong Zhu, Shaohui Lin, Xiaoyan Luo, Yunhang Shen, Xuhui Liu,
  Juan Zhang, Guodong Guo, and Baochang Zhang.
\newblock Fusion-mamba for cross-modality object detection, 2024.

\bibitem[Dosovitskiy et~al.(2020)Dosovitskiy, Beyer, Kolesnikov, Weissenborn,
  Zhai, Unterthiner, Dehghani, Minderer, Heigold, Gelly,
  et~al.]{dosovitskiy2020image}
Alexey Dosovitskiy, Lucas Beyer, Alexander Kolesnikov, Dirk Weissenborn,
  Xiaohua Zhai, Thomas Unterthiner, Mostafa Dehghani, Matthias Minderer, Georg
  Heigold, Sylvain Gelly, et~al.
\newblock An image is worth 16x16 words: Transformers for image recognition at
  scale.
\newblock \emph{arXiv preprint arXiv:2010.11929}, 2020.

\bibitem[Durbin \& Koopman(2012)Durbin and Koopman]{durbin2012time}
James Durbin and Siem~Jan Koopman.
\newblock \emph{Time series analysis by state space methods}, volume~38.
\newblock OUP Oxford, 2012.

\bibitem[Dwivedi et~al.(2022)Dwivedi, Ramp{\'a}{\v{s}}ek, Galkin, Parviz, Wolf,
  Luu, and Beaini]{dwivedi2022long}
Vijay~Prakash Dwivedi, Ladislav Ramp{\'a}{\v{s}}ek, Michael Galkin, Ali Parviz,
  Guy Wolf, Anh~Tuan Luu, and Dominique Beaini.
\newblock Long range graph benchmark.
\newblock \emph{Advances in Neural Information Processing Systems},
  35:\penalty0 22326--22340, 2022.

\bibitem[Erol et~al.(2024)Erol, Senocak, Feng, and Chung]{erol2024audio}
Mehmet~Hamza Erol, Arda Senocak, Jiu Feng, and Joon~Son Chung.
\newblock Audio mamba: Bidirectional state space model for audio representation
  learning, 2024.

\bibitem[Fathi et~al.(2023)Fathi, Pilault, Bacon, Pal, Firat, and
  Goroshin]{fathi2023block}
Mahan Fathi, Jonathan Pilault, Pierre-Luc Bacon, Christopher Pal, Orhan Firat,
  and Ross Goroshin.
\newblock Block-state transformer.
\newblock \emph{arXiv preprint arXiv:2306.09539}, 2023.

\bibitem[Fei et~al.(2024)Fei, Fan, Yu, Li, Zhang, and Huang]{fei2024dimba}
Zhengcong Fei, Mingyuan Fan, Changqian Yu, Debang Li, Youqiang Zhang, and
  Junshi Huang.
\newblock Dimba: Transformer-mamba diffusion models.
\newblock \emph{arXiv preprint arXiv:2406.01159}, 2024.

\bibitem[Fermanian et~al.(2021)Fermanian, Marion, Vert, and
  Biau]{fermanian2021framing}
Adeline Fermanian, Pierre Marion, Jean-Philippe Vert, and G{\'e}rard Biau.
\newblock Framing rnn as a kernel method: A neural ode approach.
\newblock \emph{Advances in Neural Information Processing Systems},
  34:\penalty0 3121--3134, 2021.

\bibitem[Fu et~al.(2022)Fu, Dao, Saab, Thomas, Rudra, and R{\'e}]{fu2022hungry}
Daniel~Y Fu, Tri Dao, Khaled~K Saab, Armin~W Thomas, Atri Rudra, and
  Christopher R{\'e}.
\newblock Hungry hungry hippos: Towards language modeling with state space
  models.
\newblock \emph{arXiv preprint arXiv:2212.14052}, 2022.

\bibitem[Fu et~al.(2023)Fu, Epstein, Nguyen, Thomas, Zhang, Dao, Rudra, and
  R{\'e}]{fu2023simple}
Daniel~Y Fu, Elliot~L Epstein, Eric Nguyen, Armin~W Thomas, Michael Zhang, Tri
  Dao, Atri Rudra, and Christopher R{\'e}.
\newblock Simple hardware-efficient long convolutions for sequence modeling.
\newblock In \emph{International Conference on Machine Learning}, pp.\
  10373--10391. PMLR, 2023.

\bibitem[Fu et~al.(2024)Fu, Sun, Li, Ren, Zhang, Jing, and
  Ghamisi]{fu2024hdmba}
Hang Fu, Genyun Sun, Yinhe Li, Jinchang Ren, Aizhu Zhang, Cheng Jing, and
  Pedram Ghamisi.
\newblock Hdmba: Hyperspectral remote sensing imagery dehazing with state space
  model, 2024.

\bibitem[Gao et~al.(2024)Gao, Huang, Sun, Jie, Zhong, and Ma]{gao2024matten}
Yu~Gao, Jiancheng Huang, Xiaopeng Sun, Zequn Jie, Yujie Zhong, and Lin Ma.
\newblock Matten: Video generation with mamba-attention, 2024.

\bibitem[Gilmer et~al.(2020)Gilmer, Schoenholz, Riley, Vinyals, and
  Dahl]{gilmer2020message}
Justin Gilmer, Samuel~S Schoenholz, Patrick~F Riley, Oriol Vinyals, and
  George~E Dahl.
\newblock Message passing neural networks.
\newblock \emph{Machine learning meets quantum physics}, pp.\  199--214, 2020.

\bibitem[Gong et~al.(2024)Gong, Kang, Wang, Wan, and Li]{gong2024nnmamba}
Haifan Gong, Luoyao Kang, Yitao Wang, Xiang Wan, and Haofeng Li.
\newblock nnmamba: 3d biomedical image segmentation, classification and
  landmark detection with state space model.
\newblock \emph{arXiv preprint arXiv:2402.03526}, 2024.

\bibitem[Gu \& Dao(2023)Gu and Dao]{gu2023mamba}
Albert Gu and Tri Dao.
\newblock Mamba: Linear-time sequence modeling with selective state spaces.
\newblock \emph{arXiv preprint arXiv:2312.00752}, 2023.

\bibitem[Gu et~al.(2020)Gu, Dao, Ermon, Rudra, and R{\'e}]{gu2020hippo}
Albert Gu, Tri Dao, Stefano Ermon, Atri Rudra, and Christopher R{\'e}.
\newblock Hippo: Recurrent memory with optimal polynomial projections.
\newblock \emph{Advances in neural information processing systems},
  33:\penalty0 1474--1487, 2020.

\bibitem[Gu et~al.(2021{\natexlab{a}})Gu, Goel, and R{\'e}]{gu2021efficiently}
Albert Gu, Karan Goel, and Christopher R{\'e}.
\newblock Efficiently modeling long sequences with structured state spaces.
\newblock \emph{arXiv preprint arXiv:2111.00396}, 2021{\natexlab{a}}.

\bibitem[Gu et~al.(2021{\natexlab{b}})Gu, Johnson, Goel, Saab, Dao, Rudra, and
  R{\'e}]{gu2021combining}
Albert Gu, Isys Johnson, Karan Goel, Khaled Saab, Tri Dao, Atri Rudra, and
  Christopher R{\'e}.
\newblock Combining recurrent, convolutional, and continuous-time models with
  linear state space layers.
\newblock \emph{Advances in neural information processing systems},
  34:\penalty0 572--585, 2021{\natexlab{b}}.

\bibitem[Gu et~al.(2022)Gu, Johnson, Timalsina, Rudra, and R{\'e}]{gu2022train}
Albert Gu, Isys Johnson, Aman Timalsina, Atri Rudra, and Christopher R{\'e}.
\newblock How to train your hippo: State space models with generalized
  orthogonal basis projections.
\newblock \emph{arXiv preprint arXiv:2206.12037}, 2022.

\bibitem[Guo et~al.(2024{\natexlab{a}})Guo, Li, Dai, Ouyang, Ren, and
  Xia]{guo2024mambair}
Hang Guo, Jinmin Li, Tao Dai, Zhihao Ouyang, Xudong Ren, and Shu-Tao Xia.
\newblock Mambair: A simple baseline for image restoration with state-space
  model.
\newblock \emph{arXiv preprint arXiv:2402.15648}, 2024{\natexlab{a}}.

\bibitem[Guo et~al.(2024{\natexlab{b}})Guo, Wang, and Meng]{guo2024mambamorph}
Tao Guo, Yinuo Wang, and Cai Meng.
\newblock Mambamorph: a mamba-based backbone with contrastive feature learning
  for deformable mr-ct registration.
\newblock \emph{arXiv preprint arXiv:2401.13934}, 2024{\natexlab{b}}.

\bibitem[Han et~al.(2024)Han, Tang, Wang, and Li]{han2024mamba3d}
Xu~Han, Yuan Tang, Zhaoxuan Wang, and Xianzhi Li.
\newblock Mamba3d: Enhancing local features for 3d point cloud analysis via
  state space model, 2024.

\bibitem[He et~al.(2024{\natexlab{a}})He, Han, Tang, Wang, Yang, Guo, and
  Wang]{he2024densemamba}
Wei He, Kai Han, Yehui Tang, Chengcheng Wang, Yujie Yang, Tianyu Guo, and Yunhe
  Wang.
\newblock Densemamba: State space models with dense hidden connection for
  efficient large language models.
\newblock \emph{arXiv preprint arXiv:2403.00818}, 2024{\natexlab{a}}.

\bibitem[He et~al.(2024{\natexlab{b}})He, Cao, Yan, Li, Xie, Zhang, and
  Zhou]{he2024pan}
Xuanhua He, Ke~Cao, Keyu Yan, Rui Li, Chengjun Xie, Jie Zhang, and Man Zhou.
\newblock Pan-mamba: Effective pan-sharpening with state space model.
\newblock \emph{arXiv preprint arXiv:2402.12192}, 2024{\natexlab{b}}.

\bibitem[Hu et~al.(2024)Hu, Baumann, Gui, Grebenkova, Ma, Fischer, and
  Ommer]{hu2024zigma}
Vincent~Tao Hu, Stefan~Andreas Baumann, Ming Gui, Olga Grebenkova, Pingchuan
  Ma, Johannes Fischer, and Bjorn Ommer.
\newblock Zigma: Zigzag mamba diffusion model.
\newblock \emph{arXiv preprint arXiv:2403.13802}, 2024.

\bibitem[Huang et~al.(2024{\natexlab{a}})Huang, Wang, Wang, Wu, Wang, and
  Jiang]{huang2024mambafetrack}
Ju~Huang, Shiao Wang, Shuai Wang, Zhe Wu, Xiao Wang, and Bo~Jiang.
\newblock Mamba-fetrack: Frame-event tracking via state space model,
  2024{\natexlab{a}}.

\bibitem[Huang et~al.(2024{\natexlab{b}})Huang, Pei, You, Wang, Qian, and
  Xu]{huang2024localmamba}
Tao Huang, Xiaohuan Pei, Shan You, Fei Wang, Chen Qian, and Chang Xu.
\newblock Localmamba: Visual state space model with windowed selective scan.
\newblock \emph{arXiv preprint arXiv:2403.09338}, 2024{\natexlab{b}}.

\bibitem[Jiang et~al.(2024)Jiang, Han, and Mesgarani]{jiang2024dualpath}
Xilin Jiang, Cong Han, and Nima Mesgarani.
\newblock Dual-path mamba: Short and long-term bidirectional selective
  structured state space models for speech separation, 2024.

\bibitem[Kalman(1960)]{kalman1960new}
Rudolph~Emil Kalman.
\newblock A new approach to linear filtering and prediction problems.
\newblock \emph{Journal of Basic Engineering}, 1960.

\bibitem[Katharopoulos et~al.(2020{\natexlab{a}})Katharopoulos, Vyas, Pappas,
  and Fleuret]{katharopoulos2020transformers}
Angelos Katharopoulos, Apoorv Vyas, Nikolaos Pappas, and Fran{\c{c}}ois
  Fleuret.
\newblock Transformers are rnns: Fast autoregressive transformers with linear
  attention.
\newblock In \emph{International conference on machine learning}, pp.\
  5156--5165. PMLR, 2020{\natexlab{a}}.

\bibitem[Katharopoulos et~al.(2020{\natexlab{b}})Katharopoulos, Vyas, Pappas,
  and Fleuret]{katharopoulos2020transrnn}
Angelos Katharopoulos, Apoorv Vyas, Nikolaos Pappas, and Fran{\c{c}}ois
  Fleuret.
\newblock Transformers are rnns: Fast autoregressive transformers with linear
  attention.
\newblock In \emph{International conference on machine learning}, pp.\
  5156--5165. PMLR, 2020{\natexlab{b}}.

\bibitem[Khan et~al.(2024)Khan, Asad, Benning, Roney, and
  Slabaugh]{khan2024convolution}
Abbas Khan, Muhammad Asad, Martin Benning, Caroline Roney, and Gregory
  Slabaugh.
\newblock Convolution and attention-free mamba-based cardiac image
  segmentation, 2024.

\bibitem[Koller \& Friedman(2009)Koller and Friedman]{koller2009probabilistic}
Daphne Koller and Nir Friedman.
\newblock \emph{Probabilistic graphical models: principles and techniques}.
\newblock MIT press, 2009.

\bibitem[Li et~al.(2024{\natexlab{a}})Li, Liu, Fu, Xu, and
  Zha]{li2024fouriermamba}
Dong Li, Yidi Liu, Xueyang Fu, Senyan Xu, and Zheng-Jun Zha.
\newblock Fouriermamba: Fourier learning integration with state space models
  for image deraining, 2024{\natexlab{a}}.

\bibitem[Li \& Chen(2024)Li and Chen]{li2024spmamba}
Kai Li and Guo Chen.
\newblock Spmamba: State-space model is all you need in speech separation,
  2024.

\bibitem[Li et~al.(2024{\natexlab{b}})Li, Li, Wang, He, Wang, Wang, and
  Qiao]{li2024videomamba}
Kunchang Li, Xinhao Li, Yi~Wang, Yinan He, Yali Wang, Limin Wang, and Yu~Qiao.
\newblock Videomamba: State space model for efficient video understanding.
\newblock \emph{arXiv preprint arXiv:2403.06977}, 2024{\natexlab{b}}.

\bibitem[Li et~al.(2024{\natexlab{c}})Li, Wang, Zhang, and Coster]{li2024stg}
Lincan Li, Hanchen Wang, Wenjie Zhang, and Adelle Coster.
\newblock Stg-mamba: Spatial-temporal graph learning via selective state space
  model.
\newblock \emph{arXiv preprint arXiv:2403.12418}, 2024{\natexlab{c}}.

\bibitem[Li et~al.(2024{\natexlab{d}})Li, Singh, and Grover]{li2024mamba}
Shufan Li, Harkanwar Singh, and Aditya Grover.
\newblock Mamba-nd: Selective state space modeling for multi-dimensional data.
\newblock \emph{arXiv preprint arXiv:2402.05892}, 2024{\natexlab{d}}.

\bibitem[Li et~al.(2024{\natexlab{e}})Li, Yang, and Fei]{li20243dmambacomplete}
Yixuan Li, Weidong Yang, and Ben Fei.
\newblock 3dmambacomplete: Exploring structured state space model for point
  cloud completion, 2024{\natexlab{e}}.

\bibitem[Li et~al.(2024{\natexlab{f}})Li, Pan, Zhang, Wang, and
  Yu]{li2024mambadfuse}
Zhe Li, Haiwei Pan, Kejia Zhang, Yuhua Wang, and Fengming Yu.
\newblock Mambadfuse: A mamba-based dual-phase model for multi-modality image
  fusion, 2024{\natexlab{f}}.

\bibitem[Liang et~al.(2024{\natexlab{a}})Liang, Jiang, Sun, Shi, and
  Li]{liang2024bimamba}
Aobo Liang, Xingguo Jiang, Yan Sun, Xiaohou Shi, and Ke~Li.
\newblock Bi-mamba+: Bidirectional mamba for time series forecasting,
  2024{\natexlab{a}}.

\bibitem[Liang et~al.(2024{\natexlab{b}})Liang, Zhou, Xu, Zhu, Zou, Ye, Tan,
  and Bai]{liang2024pointmamba}
Dingkang Liang, Xin Zhou, Wei Xu, Xingkui Zhu, Zhikang Zou, Xiaoqing Ye, Xiao
  Tan, and Xiang Bai.
\newblock Pointmamba: A simple state space model for point cloud analysis,
  2024{\natexlab{b}}.

\bibitem[Liao et~al.(2024)Liao, Zhu, Wang, Pan, Wang, and
  Ma]{liao2024lightmunet}
Weibin Liao, Yinghao Zhu, Xinyuan Wang, Chengwei Pan, Yasha Wang, and Liantao
  Ma.
\newblock Lightm-unet: Mamba assists in lightweight unet for medical image
  segmentation, 2024.

\bibitem[Lieber et~al.(2024)Lieber, Lenz, Bata, Cohen, Osin, Dalmedigos,
  Safahi, Meirom, Belinkov, Shalev-Shwartz, et~al.]{lieber2024jamba}
Opher Lieber, Barak Lenz, Hofit Bata, Gal Cohen, Jhonathan Osin, Itay
  Dalmedigos, Erez Safahi, Shaked Meirom, Yonatan Belinkov, Shai
  Shalev-Shwartz, et~al.
\newblock Jamba: A hybrid transformer-mamba language model.
\newblock \emph{arXiv preprint arXiv:2403.19887}, 2024.

\bibitem[Lin \& Hu(2024)Lin and Hu]{lin2024audio}
Jiaju Lin and Haoxuan Hu.
\newblock Audio mamba: Pretrained audio state space model for audio tagging,
  2024.

\bibitem[Lin et~al.(2024)Lin, Lin, Chen, and Hua]{lin2024pixmamba}
Wei-Tung Lin, Yong-Xiang Lin, Jyun-Wei Chen, and Kai-Lung Hua.
\newblock Pixmamba: Leveraging state space models in a dual-level architecture
  for underwater image enhancement, 2024.

\bibitem[Liu et~al.(2024{\natexlab{a}})Liu, Lin, Wang, Liu, and
  Caverlee]{liu2024mamba4rec}
Chengkai Liu, Jianghao Lin, Jianling Wang, Hanzhou Liu, and James Caverlee.
\newblock Mamba4rec: Towards efficient sequential recommendation with selective
  state space models.
\newblock \emph{arXiv preprint arXiv:2403.03900}, 2024{\natexlab{a}}.

\bibitem[Liu et~al.(2024{\natexlab{b}})Liu, Liu, Wang, Lee, Zhou, An, Yang,
  Zhang, Guo, and Zhang]{liu2024robomamba}
Jiaming Liu, Mengzhen Liu, Zhenyu Wang, Lily Lee, Kaichen Zhou, Pengju An,
  Senqiao Yang, Renrui Zhang, Yandong Guo, and Shanghang Zhang.
\newblock Robomamba: Multimodal state space model for efficient robot reasoning
  and manipulation, 2024{\natexlab{b}}.

\bibitem[Liu et~al.(2024{\natexlab{c}})Liu, Yang, Zhou, Xi, Yu, Yu, Liang, Shi,
  Zhang, Zheng, et~al.]{liu2024swin}
Jiarun Liu, Hao Yang, Hong-Yu Zhou, Yan Xi, Lequan Yu, Yizhou Yu, Yong Liang,
  Guangming Shi, Shaoting Zhang, Hairong Zheng, et~al.
\newblock Swin-umamba: Mamba-based unet with imagenet-based pretraining.
\newblock \emph{arXiv preprint arXiv:2402.03302}, 2024{\natexlab{c}}.

\bibitem[Liu et~al.(2024{\natexlab{d}})Liu, Han, Liu, Aviles-Rivero, Jiang,
  Liu, and Wang]{liu2024mamba4d}
Jiuming Liu, Jinru Han, Lihao Liu, Angelica~I. Aviles-Rivero, Chaokang Jiang,
  Zhe Liu, and Hesheng Wang.
\newblock Mamba4d: Efficient long-sequence point cloud video understanding with
  disentangled spatial-temporal state space models, 2024{\natexlab{d}}.

\bibitem[Liu et~al.(2024{\natexlab{e}})Liu, Yu, Wang, Zheng, Deng, Ye, and
  Wang]{liu2024point}
Jiuming Liu, Ruiji Yu, Yian Wang, Yu~Zheng, Tianchen Deng, Weicai Ye, and
  Hesheng Wang.
\newblock Point mamba: A novel point cloud backbone based on state space model
  with octree-based ordering strategy.
\newblock \emph{arXiv preprint arXiv:2403.06467}, 2024{\natexlab{e}}.

\bibitem[Liu et~al.(2024{\natexlab{f}})Liu, Dan, Lu, Yu, Li, and
  Li]{liu2024cmunet}
Mushui Liu, Jun Dan, Ziqian Lu, Yunlong Yu, Yingming Li, and Xi~Li.
\newblock Cm-unet: Hybrid cnn-mamba unet for remote sensing image semantic
  segmentation, 2024{\natexlab{f}}.

\bibitem[Liu et~al.(2024{\natexlab{g}})Liu, Tian, Zhao, Yu, Xie, Wang, Ye, and
  Liu]{liu2024vmamba}
Yue Liu, Yunjie Tian, Yuzhong Zhao, Hongtian Yu, Lingxi Xie, Yaowei Wang,
  Qixiang Ye, and Yunfan Liu.
\newblock Vmamba: Visual state space model.
\newblock \emph{arXiv preprint arXiv:2401.10166}, 2024{\natexlab{g}}.

\bibitem[Liu et~al.(2021)Liu, Lin, Cao, Hu, Wei, Zhang, Lin, and
  Guo]{liu2021swin}
Ze~Liu, Yutong Lin, Yue Cao, Han Hu, Yixuan Wei, Zheng Zhang, Stephen Lin, and
  Baining Guo.
\newblock Swin transformer: Hierarchical vision transformer using shifted
  windows.
\newblock In \emph{Proceedings of the IEEE/CVF international conference on
  computer vision}, pp.\  10012--10022, 2021.

\bibitem[Ma et~al.(2024{\natexlab{a}})Ma, Lei, Celik, and
  Li]{ma2024feryolomamba}
Hui Ma, Sen Lei, Turgay Celik, and Heng-Chao Li.
\newblock Fer-yolo-mamba: Facial expression detection and classification based
  on selective state space, 2024{\natexlab{a}}.

\bibitem[Ma et~al.(2024{\natexlab{b}})Ma, Li, and Wang]{ma2024u}
Jun Ma, Feifei Li, and Bo~Wang.
\newblock U-mamba: Enhancing long-range dependency for biomedical image
  segmentation.
\newblock \emph{arXiv preprint arXiv:2401.04722}, 2024{\natexlab{b}}.

\bibitem[Ma et~al.(2024{\natexlab{c}})Ma, Zhang, and Pun]{ma2024rs3mamba}
Xianping Ma, Xiaokang Zhang, and Man-On Pun.
\newblock Rs3mamba: Visual state space model for remote sensing images semantic
  segmentation, 2024{\natexlab{c}}.

\bibitem[Mo(2024)]{mo2024efficient}
Shentong Mo.
\newblock Efficient 3d shape generation via diffusion mamba with bidirectional
  ssms, 2024.

\bibitem[Mo \& Tian(2024)Mo and Tian]{mo2024scaling}
Shentong Mo and Yapeng Tian.
\newblock Scaling diffusion mamba with bidirectional ssms for efficient image
  and video generation, 2024.

\bibitem[Park et~al.(2024)Park, Park, Xiong, Lee, Cho, Oymak, Lee, and
  Papailiopoulos]{park2024can}
Jongho Park, Jaeseung Park, Zheyang Xiong, Nayoung Lee, Jaewoong Cho, Samet
  Oymak, Kangwook Lee, and Dimitris Papailiopoulos.
\newblock Can mamba learn how to learn? a comparative study on in-context
  learning tasks.
\newblock \emph{arXiv preprint arXiv:2402.04248}, 2024.

\bibitem[Patro \& Agneeswaran(2024{\natexlab{a}})Patro and
  Agneeswaran]{patro2024simba}
Badri~N. Patro and Vijay~S. Agneeswaran.
\newblock Simba: Simplified mamba-based architecture for vision and
  multivariate time series, 2024{\natexlab{a}}.

\bibitem[Patro et~al.(2024)Patro, Ranganath, Namboodiri, and
  Agneeswaran]{Heracles}
Badri~N. Patro, Suhas Ranganath, Vinay~P. Namboodiri, and Vijay~Srinivas
  Agneeswaran.
\newblock Heracles: A hybrid ssm-transformer model for high-resolution image
  and time-series analysis.
\newblock \emph{arXiv preprint arXiv:2403.18063v2}, 2024.

\bibitem[Patro \& Agneeswaran(2024{\natexlab{b}})Patro and
  Agneeswaran]{patro2024mamba360}
Badri~Narayana Patro and Vijay~Srinivas Agneeswaran.
\newblock Mamba-360: Survey of state space models as transformer alternative
  for long sequence modelling: Methods, applications, and challenges,
  2024{\natexlab{b}}.

\bibitem[Pei et~al.(2024)Pei, Huang, and Xu]{pei2024efficientvmamba}
Xiaohuan Pei, Tao Huang, and Chang Xu.
\newblock Efficientvmamba: Atrous selective scan for light weight visual mamba.
\newblock \emph{arXiv preprint arXiv:2403.09977}, 2024.

\bibitem[Peng et~al.(2024)Peng, Zhu, Deng, Lei, and Deng]{peng2024fusionmamba}
Siran Peng, Xiangyu Zhu, Haoyu Deng, Zhen Lei, and Liang-Jian Deng.
\newblock Fusionmamba: Efficient image fusion with state space model.
\newblock \emph{arXiv preprint arXiv:2404.07932}, 2024.

\bibitem[Pi{\'o}ro et~al.(2024)Pi{\'o}ro, Ciebiera, Kr{\'o}l, Ludziejewski, and
  Jaszczur]{pioro2024moe}
Maciej Pi{\'o}ro, Kamil Ciebiera, Krystian Kr{\'o}l, Jan Ludziejewski, and
  Sebastian Jaszczur.
\newblock Moe-mamba: Efficient selective state space models with mixture of
  experts.
\newblock \emph{arXiv preprint arXiv:2401.04081}, 2024.

\bibitem[Poli et~al.(2023)Poli, Massaroli, Nguyen, Fu, Dao, Baccus, Bengio,
  Ermon, and R{\'e}]{poli2023hyena}
Michael Poli, Stefano Massaroli, Eric Nguyen, Daniel~Y Fu, Tri Dao, Stephen
  Baccus, Yoshua Bengio, Stefano Ermon, and Christopher R{\'e}.
\newblock Hyena hierarchy: Towards larger convolutional language models.
\newblock \emph{arXiv preprint arXiv:2302.10866}, 2023.

\bibitem[Qian et~al.(2024)Qian, Xiao, Wu, Yang, Li, Wang, Wang, Kou, and
  Zhang]{qian2024smcd}
Ziyun Qian, Zeyu Xiao, Zhenyi Wu, Dingkang Yang, Mingcheng Li, Shunli Wang,
  Shuaibing Wang, Dongliang Kou, and Lihua Zhang.
\newblock Smcd: High realism motion style transfer via mamba-based diffusion,
  2024.

\bibitem[Qiao et~al.(2024)Qiao, Yu, Guo, Chen, Zhao, Sun, Wu, and
  Liu]{qiao2024vl}
Yanyuan Qiao, Zheng Yu, Longteng Guo, Sihan Chen, Zijia Zhao, Mingzhen Sun,
  Qi~Wu, and Jing Liu.
\newblock Vl-mamba: Exploring state space models for multimodal learning.
\newblock \emph{arXiv preprint arXiv:2403.13600}, 2024.

\bibitem[Quan \& Li(2024)Quan and Li]{quan2024multichannel}
Changsheng Quan and Xiaofei Li.
\newblock Multichannel long-term streaming neural speech enhancement for static
  and moving speakers.
\newblock \emph{arXiv preprint arXiv:2403.07675}, 2024.

\bibitem[Radford et~al.(2018)Radford, Narasimhan, Salimans, Sutskever,
  et~al.]{radford2018improving}
Alec Radford, Karthik Narasimhan, Tim Salimans, Ilya Sutskever, et~al.
\newblock \emph{Improving language understanding by generative pre-training}.
\newblock OpenAI, 2018.

\bibitem[Ramp{\'a}{\v{s}}ek et~al.(2022)Ramp{\'a}{\v{s}}ek, Galkin, Dwivedi,
  Luu, Wolf, and Beaini]{rampavsek2022recipe}
Ladislav Ramp{\'a}{\v{s}}ek, Michael Galkin, Vijay~Prakash Dwivedi, Anh~Tuan
  Luu, Guy Wolf, and Dominique Beaini.
\newblock Recipe for a general, powerful, scalable graph transformer.
\newblock \emph{Advances in Neural Information Processing Systems},
  35:\penalty0 14501--14515, 2022.

\bibitem[Ren et~al.(2024)Ren, Liu, Lu, Shen, Liang, and Chen]{ren2024samba}
Liliang Ren, Yang Liu, Yadong Lu, Yelong Shen, Chen Liang, and Weizhu Chen.
\newblock Samba: Simple hybrid state space models for efficient unlimited
  context language modeling, 2024.

\bibitem[Ronneberger et~al.(2015)Ronneberger, Fischer, and
  Brox]{ronneberger2015u}
Olaf Ronneberger, Philipp Fischer, and Thomas Brox.
\newblock U-net: Convolutional networks for biomedical image segmentation.
\newblock In \emph{Medical Image Computing and Computer-Assisted
  Intervention--MICCAI 2015: 18th International Conference, Munich, Germany,
  October 5-9, 2015, Proceedings, Part III 18}, pp.\  234--241. Springer, 2015.

\bibitem[Ruan \& Xiang(2024)Ruan and Xiang]{ruan2024vm}
Jiacheng Ruan and Suncheng Xiang.
\newblock Vm-unet: Vision mamba unet for medical image segmentation.
\newblock \emph{arXiv preprint arXiv:2402.02491}, 2024.

\bibitem[Rusch et~al.(2023)Rusch, Bronstein, and Mishra]{rusch2023survey}
T~Konstantin Rusch, Michael~M Bronstein, and Siddhartha Mishra.
\newblock A survey on oversmoothing in graph neural networks.
\newblock \emph{arXiv preprint arXiv:2303.10993}, 2023.

\bibitem[Russakovsky et~al.(2015)Russakovsky, Deng, Su, Krause, Satheesh, Ma,
  Huang, Karpathy, Khosla, Bernstein, Berg, and Fei-Fei]{imagenet15russakovsky}
Olga Russakovsky, Jia Deng, Hao Su, Jonathan Krause, Sanjeev Satheesh, Sean Ma,
  Zhiheng Huang, Andrej Karpathy, Aditya Khosla, Michael Bernstein,
  Alexander~C. Berg, and Li~Fei-Fei.
\newblock {ImageNet Large Scale Visual Recognition Challenge}.
\newblock \emph{International Journal of Computer Vision (IJCV)}, 115\penalty0
  (3):\penalty0 211--252, 2015.
\newblock \doi{10.1007/s11263-015-0816-y}.

\bibitem[Schiff et~al.(2024)Schiff, Kao, Gokaslan, Dao, Gu, and
  Kuleshov]{schiff2024caduceus}
Yair Schiff, Chia-Hsiang Kao, Aaron Gokaslan, Tri Dao, Albert Gu, and Volodymyr
  Kuleshov.
\newblock Caduceus: Bi-directional equivariant long-range dna sequence
  modeling.
\newblock \emph{arXiv preprint arXiv:2403.03234}, 2024.

\bibitem[Shams et~al.(2024)Shams, Dindar, Jiang, and
  Mesgarani]{shams2024ssamba}
Siavash Shams, Sukru~Samet Dindar, Xilin Jiang, and Nima Mesgarani.
\newblock Ssamba: Self-supervised audio representation learning with mamba
  state space model, 2024.

\bibitem[Shen et~al.(2024)Shen, Wu, Yi, Zhou, Zhang, Yan, and
  Wang]{shen2024gamba}
Qiuhong Shen, Zike Wu, Xuanyu Yi, Pan Zhou, Hanwang Zhang, Shuicheng Yan, and
  Xinchao Wang.
\newblock Gamba: Marry gaussian splatting with mamba for single view 3d
  reconstruction, 2024.

\bibitem[Sheng et~al.(2024)Sheng, Zhou, Wang, Ye, and Fan]{sheng2024dualmamba}
Jiamu Sheng, Jingyi Zhou, Jiong Wang, Peng Ye, and Jiayuan Fan.
\newblock Dualmamba: A lightweight spectral-spatial mamba-convolution network
  for hyperspectral image classification, 2024.

\bibitem[Shi et~al.(2023)Shi, Chen, Misra, Scales, Dohan, Chi, Sch{\"a}rli, and
  Zhou]{shi2023large}
Freda Shi, Xinyun Chen, Kanishka Misra, Nathan Scales, David Dohan, Ed~H Chi,
  Nathanael Sch{\"a}rli, and Denny Zhou.
\newblock Large language models can be easily distracted by irrelevant context.
\newblock In \emph{International Conference on Machine Learning}, pp.\
  31210--31227. PMLR, 2023.

\bibitem[Shi(2024)]{shi2024mambastock}
Zhuangwei Shi.
\newblock Mambastock: Selective state space model for stock prediction.
\newblock \emph{arXiv preprint arXiv:2402.18959}, 2024.

\bibitem[Tang et~al.(2024)Tang, Cheng, Huang, Tan, Lu, and Wu]{tang2024rotate}
Hao Tang, Lianglun Cheng, Guoheng Huang, Zhengguang Tan, Junhao Lu, and Kaihong
  Wu.
\newblock Rotate to scan: Unet-like mamba with triplet ssm module for medical
  image segmentation, 2024.

\bibitem[Teng et~al.(2024)Teng, Wu, Shi, Ning, Dai, Wang, Li, and
  Liu]{teng2024dim}
Yao Teng, Yue Wu, Han Shi, Xuefei Ning, Guohao Dai, Yu~Wang, Zhenguo Li, and
  Xihui Liu.
\newblock Dim: Diffusion mamba for efficient high-resolution image synthesis.
\newblock \emph{arXiv preprint arXiv:2405.14224}, 2024.

\bibitem[Tsai et~al.(2024)Tsai, Lin, Hu, Chang, Zhu, and Wang]{tsai2024uumamba}
Ting~Yu Tsai, Li~Lin, Shu Hu, Ming-Ching Chang, Hongtu Zhu, and Xin Wang.
\newblock Uu-mamba: Uncertainty-aware u-mamba for cardiac image segmentation,
  2024.

\bibitem[Tsai et~al.(2019)Tsai, Bai, Yamada, Morency, and
  Salakhutdinov]{tsai2019transformer}
Yao-Hung~Hubert Tsai, Shaojie Bai, Makoto Yamada, Louis-Philippe Morency, and
  Ruslan Salakhutdinov.
\newblock Transformer dissection: a unified understanding of transformer's
  attention via the lens of kernel.
\newblock \emph{arXiv preprint arXiv:1908.11775}, 2019.

\bibitem[Vaswani et~al.(2017)Vaswani, Shazeer, Parmar, Uszkoreit, Jones, Gomez,
  Kaiser, and Polosukhin]{vaswani2017attention}
Ashish Vaswani, Noam Shazeer, Niki Parmar, Jakob Uszkoreit, Llion Jones,
  Aidan~N Gomez, {\L}ukasz Kaiser, and Illia Polosukhin.
\newblock Attention is all you need.
\newblock \emph{Advances in neural information processing systems}, 30, 2017.

\bibitem[Verma et~al.(2024)Verma, Singh, Bhartari, Jarwal, Singh, and
  Singh]{verma2024soar}
Tushar Verma, Jyotsna Singh, Yash Bhartari, Rishi Jarwal, Suraj Singh, and
  Shubhkarman Singh.
\newblock Soar: Advancements in small body object detection for aerial imagery
  using state space models and programmable gradients, 2024.

\bibitem[Wan et~al.(2024)Wan, Wang, Yong, Zhang, Stepputtis, Sycara, and
  Xie]{wan2024sigma}
Zifu Wan, Yuhao Wang, Silong Yong, Pingping Zhang, Simon Stepputtis, Katia
  Sycara, and Yaqi Xie.
\newblock Sigma: Siamese mamba network for multi-modal semantic segmentation,
  2024.

\bibitem[Wang et~al.(2024{\natexlab{a}})Wang, Tsepa, Ma, and
  Wang]{wang2024graph-mamba}
Chloe Wang, Oleksii Tsepa, Jun Ma, and Bo~Wang.
\newblock Graph-mamba: Towards long-range graph sequence modeling with
  selective state spaces.
\newblock \emph{arXiv preprint arXiv:2402.00789}, 2024{\natexlab{a}}.

\bibitem[Wang et~al.(2022)Wang, Yan, Gu, and Rush]{wang2022pretraining}
Junxiong Wang, Jing~Nathan Yan, Albert Gu, and Alexander~M Rush.
\newblock Pretraining without attention.
\newblock \emph{arXiv preprint arXiv:2212.10544}, 2022.

\bibitem[Wang et~al.(2024{\natexlab{b}})Wang, Gangavarapu, Yan, and
  Rush]{wang2024mambabyte}
Junxiong Wang, Tushaar Gangavarapu, Jing~Nathan Yan, and Alexander~M Rush.
\newblock Mambabyte: Token-free selective state space model.
\newblock \emph{arXiv preprint arXiv:2401.13660}, 2024{\natexlab{b}}.

\bibitem[Wang et~al.(2024{\natexlab{c}})Wang, Xie, Wang, Wang, Zhao, and
  Cui]{wang2024netmamba}
Tongze Wang, Xiaohui Xie, Wenduo Wang, Chuyi Wang, Youjian Zhao, and Yong Cui.
\newblock Netmamba: Efficient network traffic classification via pre-training
  unidirectional mamba, 2024{\natexlab{c}}.

\bibitem[Wang et~al.(2024{\natexlab{d}})Wang, He, and
  Zhu]{wang2024echomamba4rec}
Yuda Wang, Xuxin He, and Shengxin Zhu.
\newblock Echomamba4rec: Harmonizing bidirectional state space models with
  spectral filtering for advanced sequential recommendation,
  2024{\natexlab{d}}.

\bibitem[Wang et~al.(2024{\natexlab{e}})Wang, Li, Xu, and
  Zhu]{wang2024mambayolo}
Zeyu Wang, Chen Li, Huiying Xu, and Xinzhong Zhu.
\newblock Mamba yolo: Ssms-based yolo for object detection, 2024{\natexlab{e}}.

\bibitem[Wang et~al.(2024{\natexlab{f}})Wang, Chen, Wu, Zhao, Zhou, and
  Xu]{wang2024pointramba}
Zicheng Wang, Zhenghao Chen, Yiming Wu, Zhen Zhao, Luping Zhou, and Dong Xu.
\newblock Pointramba: A hybrid transformer-mamba framework for point cloud
  analysis, 2024{\natexlab{f}}.

\bibitem[Wang et~al.(2024{\natexlab{g}})Wang, Kong, Feng, Wang, Zhao, Wang, and
  Zhang]{wang2024Time}
Zihan Wang, Fanheng Kong, Shi Feng, Ming Wang, Han Zhao, Daling Wang, and Yifei
  Zhang.
\newblock Is mamba effective for time series forecasting?
\newblock \emph{arXiv preprint arXiv:2403.11144}, 2024{\natexlab{g}}.

\bibitem[Wang \& Ma(2024{\natexlab{a}})Wang and Ma]{wang2024semi}
Ziyang Wang and Chao Ma.
\newblock Semi-mamba-unet: Pixel-level contrastive cross-supervised visual
  mamba-based unet for semi-supervised medical image segmentation.
\newblock \emph{arXiv preprint arXiv:2402.07245}, 2024{\natexlab{a}}.

\bibitem[Wang \& Ma(2024{\natexlab{b}})Wang and Ma]{wang2024weak}
Ziyang Wang and Chao Ma.
\newblock Weak-mamba-unet: Visual mamba makes cnn and vit work better for
  scribble-based medical image segmentation.
\newblock \emph{arXiv preprint arXiv:2402.10887}, 2024{\natexlab{b}}.

\bibitem[Wang et~al.(2024{\natexlab{h}})Wang, Zheng, Zhang, Cui, and
  Li]{wang2024mamba}
Ziyang Wang, Jian-Qing Zheng, Yichi Zhang, Ge~Cui, and Lei Li.
\newblock Mamba-unet: Unet-like pure visual mamba for medical image
  segmentation.
\newblock \emph{arXiv preprint arXiv:2402.05079}, 2024{\natexlab{h}}.

\bibitem[wei Chen et~al.(2024)wei Chen, jie Xiong, and bin
  Gao]{chen2024pointabmintegrating}
Jia wei Chen, Yu~jie Xiong, and Yong bin Gao.
\newblock Pointabm:integrating bidirectional state space model with multi-head
  self-attention for point cloud analysis, 2024.

\bibitem[Wu et~al.(2024)Wu, Liu, Liang, and Chang]{wu2024h}
Renkai Wu, Yinghao Liu, Pengchen Liang, and Qing Chang.
\newblock H-vmunet: High-order vision mamba unet for medical image
  segmentation.
\newblock \emph{arXiv preprint arXiv:2403.13642}, 2024.

\bibitem[xia et~al.(2024)xia, Lu, Wang, Wang, Xia, and Zhou]{xia2024lfmamba}
Wang xia, Yao Lu, Shunzhou Wang, Ziqi Wang, Peiqi Xia, and Tianfei Zhou.
\newblock Lfmamba: Light field image super-resolution with state space model,
  2024.

\bibitem[Xie et~al.(2024)Xie, Liao, Zhang, Yi, Zhu, and Luo]{xie2024promamba}
Jianhao Xie, Ruofan Liao, Ziang Zhang, Sida Yi, Yuesheng Zhu, and Guibo Luo.
\newblock Promamba: Prompt-mamba for polyp segmentation, 2024.

\bibitem[Xing et~al.(2024)Xing, Ye, Yang, Liu, and Zhu]{xing2024segmamba}
Zhaohu Xing, Tian Ye, Yijun Yang, Guang Liu, and Lei Zhu.
\newblock Segmamba: Long-range sequential modeling mamba for 3d medical image
  segmentation.
\newblock \emph{arXiv preprint arXiv:2401.13560}, 2024.

\bibitem[Xu et~al.(2024{\natexlab{a}})Xu, Liang, Huang, Lan, and
  Shu]{xu2024integrating}
Xiongxiao Xu, Yueqing Liang, Baixiang Huang, Zhiling Lan, and Kai Shu.
\newblock Integrating mamba and transformer for long-short range time series
  forecasting.
\newblock \emph{arXiv preprint arXiv:2404.14757}, 2024{\natexlab{a}}.

\bibitem[Xu et~al.(2024{\natexlab{b}})Xu, Lin, Han, Yang, Li, Zhang, and
  Li]{xu2024mambatalk}
Zunnan Xu, Yukang Lin, Haonan Han, Sicheng Yang, Ronghui Li, Yachao Zhang, and
  Xiu Li.
\newblock Mambatalk: Efficient holistic gesture synthesis with selective state
  space models.
\newblock \emph{arXiv preprint arXiv:2403.09471}, 2024{\natexlab{b}}.

\bibitem[Yang et~al.(2024{\natexlab{a}})Yang, Du, Yang, Du, Zheng, and
  Wang]{yang2024cmvim}
Guangqian Yang, Kangrui Du, Zhihan Yang, Ye~Du, Yongping Zheng, and Shujun
  Wang.
\newblock Cmvim: Contrastive masked vim autoencoder for 3d multi-modal
  representation learning for ad classification, 2024{\natexlab{a}}.

\bibitem[Yang et~al.(2024{\natexlab{b}})Yang, Xing, and Zhu]{yang2024vivim}
Yijun Yang, Zhaohu Xing, and Lei Zhu.
\newblock Vivim: a video vision mamba for medical video object segmentation.
\newblock \emph{arXiv preprint arXiv:2401.14168}, 2024{\natexlab{b}}.

\bibitem[Yang et~al.(2024{\natexlab{c}})Yang, Ma, Yao, Zhong, Zhang, and
  Wang]{yang2024remamber}
Yuhuan Yang, Chaofan Ma, Jiangchao Yao, Zhun Zhong, Ya~Zhang, and Yanfeng Wang.
\newblock Remamber: Referring image segmentation with mamba twister,
  2024{\natexlab{c}}.

\bibitem[Yang et~al.(2024{\natexlab{d}})Yang, Mitra, Kwon, and
  Yu]{yang2024clinicalmamba}
Zhichao Yang, Avijit Mitra, Sunjae Kwon, and Hong Yu.
\newblock Clinicalmamba: A generative clinical language model on longitudinal
  clinical notes.
\newblock \emph{arXiv preprint arXiv:2403.05795}, 2024{\natexlab{d}}.

\bibitem[Yao et~al.(2024)Yao, Hong, Li, and Chanussot]{yao2024spectralmamba}
Jing Yao, Danfeng Hong, Chenyu Li, and Jocelyn Chanussot.
\newblock Spectralmamba: Efficient mamba for hyperspectral image
  classification, 2024.

\bibitem[Yu \& Wang(2024)Yu and Wang]{yu2024mambaout}
Weihao Yu and Xinchao Wang.
\newblock Mambaout: Do we really need mamba for vision?
\newblock \emph{arXiv preprint arXiv:2405.07992}, 2024.

\bibitem[Yuan et~al.(2024)Yuan, Zhao, and Agaian]{yuan2024mucmnet}
Chunyu Yuan, Dongfang Zhao, and Sos~S. Agaian.
\newblock Mucm-net: A mamba powered ucm-net for skin lesion segmentation, 2024.

\bibitem[Yue \& Li(2024)Yue and Li]{yue2024medmamba}
Yubiao Yue and Zhenzhang Li.
\newblock Medmamba: Vision mamba for medical image classification.
\newblock \emph{arXiv preprint arXiv:2403.03849}, 2024.

\bibitem[Yun et~al.(2019)Yun, Jeong, Kim, Kang, and Kim]{yun2019graph}
Seongjun Yun, Minbyul Jeong, Raehyun Kim, Jaewoo Kang, and Hyunwoo~J Kim.
\newblock Graph transformer networks.
\newblock \emph{Advances in neural information processing systems}, 32, 2019.

\bibitem[Zeng et~al.(2024)Zeng, Liu, Zheng, and Kong]{zeng2024cmamba}
Chaolv Zeng, Zhanyu Liu, Guanjie Zheng, and Linghe Kong.
\newblock C-mamba: Channel correlation enhanced state space models for
  multivariate time series forecasting, 2024.

\bibitem[Zhang et~al.(2024{\natexlab{a}})Zhang, Chen, Liu, Chen, Zou, and
  Shi]{zhang2024cdmamba}
Haotian Zhang, Keyan Chen, Chenyang Liu, Hao Chen, Zhengxia Zou, and Zhenwei
  Shi.
\newblock Cdmamba: Remote sensing image change detection with mamba.
\newblock \emph{arXiv preprint arXiv:2406.04207}, 2024{\natexlab{a}}.

\bibitem[Zhang et~al.(2024{\natexlab{b}})Zhang, Yu, Gu, Lin, and
  Tao]{zhang2024vm}
Mingya Zhang, Yue Yu, Limei Gu, Tingsheng Lin, and Xianping Tao.
\newblock Vm-unet-v2 rethinking vision mamba unet for medical image
  segmentation.
\newblock \emph{arXiv preprint arXiv:2403.09157}, 2024{\natexlab{b}}.

\bibitem[Zhang et~al.(2024{\natexlab{c}})Zhang, Li, Yuan, Ji, and
  Yan]{zhang2024point}
Tao Zhang, Xiangtai Li, Haobo Yuan, Shunping Ji, and Shuicheng Yan.
\newblock Point cloud mamba: Point cloud learning via state space model,
  2024{\natexlab{c}}.

\bibitem[Zhang et~al.(2024{\natexlab{d}})Zhang, Zhang, Liu, Xiao, Qian, Ahmed,
  Ambikairajah, Li, and Epps]{zhang2024mambaspeech}
Xiangyu Zhang, Qiquan Zhang, Hexin Liu, Tianyi Xiao, Xinyuan Qian, Beena Ahmed,
  Eliathamby Ambikairajah, Haizhou Li, and Julien Epps.
\newblock Mamba in speech: Towards an alternative to self-attention,
  2024{\natexlab{d}}.

\bibitem[Zhang et~al.(2024{\natexlab{e}})Zhang, Zeng, Pan, Shen, and
  Chen]{zhang2024llemamba}
Xuanqi Zhang, Haijin Zeng, Jinwang Pan, Qiangqiang Shen, and Yongyong Chen.
\newblock Llemamba: Low-light enhancement via relighting-guided mamba with deep
  unfolding network, 2024{\natexlab{e}}.

\bibitem[Zhang et~al.(2024{\natexlab{f}})Zhang, Liu, Reid, Hartley, Zhuang, and
  Tang]{zhang2024motion}
Zeyu Zhang, Akide Liu, Ian Reid, Richard Hartley, Bohan Zhuang, and Hao Tang.
\newblock Motion mamba: Efficient and long sequence motion generation with
  hierarchical and bidirectional selective ssm.
\newblock \emph{arXiv preprint arXiv:2403.07487}, 2024{\natexlab{f}}.

\bibitem[Zhao et~al.(2024)Zhao, Zhang, Zhao, Ding, Huang, and
  Wang]{zhao2024cobra}
Han Zhao, Min Zhang, Wei Zhao, Pengxiang Ding, Siteng Huang, and Donglin Wang.
\newblock Cobra: Extending mamba to multi-modal large language model for
  efficient inference.
\newblock \emph{arXiv preprint arXiv:2403.14520}, 2024.

\bibitem[Zhen et~al.(2024)Zhen, Hu, and Feng]{zhen2024freqmamba}
Zou Zhen, Yu~Hu, and Zhao Feng.
\newblock Freqmamba: Viewing mamba from a frequency perspective for image
  deraining, 2024.

\bibitem[Zheng \& Wu(2024)Zheng and Wu]{zheng2024u}
Zhuoran Zheng and Chen Wu.
\newblock U-shaped vision mamba for single image dehazing.
\newblock \emph{arXiv preprint arXiv:2402.04139}, 2024.

\bibitem[Zheng \& Zhang(2024)Zheng and Zhang]{zheng2024fd}
Zhuoran Zheng and Jun Zhang.
\newblock Fd-vision mamba for endoscopic exposure correction.
\newblock \emph{arXiv preprint arXiv:2402.06378}, 2024.

\bibitem[Zhou et~al.(2024{\natexlab{a}})Zhou, Wu, Chen, Chen, and
  He]{zhou2024rsdehamba}
Huiling Zhou, Xianhao Wu, Hongming Chen, Xiang Chen, and Xin He.
\newblock Rsdehamba: Lightweight vision mamba for remote sensing satellite
  image dehazing, 2024{\natexlab{a}}.

\bibitem[Zhou et~al.(2024{\natexlab{b}})Zhou, Jiang, Wu, Zhu, Wang, and
  Jin]{zhou2024mgi}
Jiaying Zhou, Mingzhou Jiang, Junde Wu, Jiayuan Zhu, Ziyue Wang, and Yueming
  Jin.
\newblock Mgi: Multimodal contrastive pre-training of genomic and medical
  imaging, 2024{\natexlab{b}}.

\bibitem[Zhou et~al.(2024{\natexlab{c}})Zhou, Yang, Fei, Xu, Zhang, Liu, Luo,
  and He]{zhou20243dmambaipf}
Qingyuan Zhou, Weidong Yang, Ben Fei, Jingyi Xu, Rui Zhang, Keyi Liu, Yeqi Luo,
  and Ying He.
\newblock 3dmambaipf: A state space model for iterative point cloud filtering
  via differentiable rendering, 2024{\natexlab{c}}.

\bibitem[Zhou et~al.(2024{\natexlab{d}})Zhou, Kamata, Wang, Wong, Huiying, and
  Hou]{zhou2024mambainmamba}
Weilian Zhou, Sei-Ichiro Kamata, Haipeng Wang, Man-Sing Wong, Huiying, and Hou.
\newblock Mamba-in-mamba: Centralized mamba-cross-scan in tokenized mamba model
  for hyperspectral image classification, 2024{\natexlab{d}}.

\bibitem[Zhu et~al.(2024{\natexlab{a}})Zhu, Huang, Liao, Liew, Yan, Feng, and
  Wang]{zhu2024dig}
Lianghui Zhu, Zilong Huang, Bencheng Liao, Jun~Hao Liew, Hanshu Yan, Jiashi
  Feng, and Xinggang Wang.
\newblock Dig: Scalable and efficient diffusion models with gated linear
  attention, 2024{\natexlab{a}}.

\bibitem[Zhu et~al.(2024{\natexlab{b}})Zhu, Liao, Zhang, Wang, Liu, and
  Wang]{zhu2024vision}
Lianghui Zhu, Bencheng Liao, Qian Zhang, Xinlong Wang, Wenyu Liu, and Xinggang
  Wang.
\newblock Vision mamba: Efficient visual representation learning with
  bidirectional state space model.
\newblock \emph{arXiv preprint arXiv:2401.09417}, 2024{\natexlab{b}}.

\bibitem[Zuo et~al.(2022)Zuo, Liu, Jiao, Charles, Manavoglu, Zhao, and
  Gao]{zuo2022efficient}
Simiao Zuo, Xiaodong Liu, Jian Jiao, Denis Charles, Eren Manavoglu, Tuo Zhao,
  and Jianfeng Gao.
\newblock Efficient long sequence modeling via state space augmented
  transformer.
\newblock \emph{arXiv preprint arXiv:2212.08136}, 2022.

\end{thebibliography}
\bibliographystyle{tmlr}




\end{document}